\documentclass[manuscript,screen]{acmart}
\AtBeginDocument{%
  \providecommand\BibTeX{{%
    \normalfont B\kern-0.5em{\scshape i\kern-0.25em b}\kern-0.8em\TeX}}}

\setcopyright{acmcopyright}
\copyrightyear{2020}
\acmYear{2020}
\acmDOI{10.1145/0000001.0000001} 

\usepackage{tikz}
\usepackage{lscape}
\usepackage{caption}
\usepackage{subcaption}

\usetikzlibrary{positioning,shapes,shadows,arrows}
\usepackage{bbm}
\usepackage{array}
\usepackage{multirow}
\usepackage{longtable}
\usepackage{rotating, float, caption}
\newcolumntype{L}[1]{>{\raggedright\let\newline\\\arraybackslash\hspace{0pt}}m{#1}}
\newcolumntype{C}[1]{>{\centering\let\newline\\\arraybackslash\hspace{0pt}}m{#1}}
\newcolumntype{R}[1]{>{\raggedleft\let\newline\\\arraybackslash\hspace{0pt}}m{#1}}

\newcolumntype{x}[1]{
>{\centering\hspace{0pt}}p{#1}}%

\begin{document}

\title{A Survey of Evaluation Metrics Used for NLG Systems}

\author{Ananya B. Sai}
\affiliation{
\institution{Robert-Bosch Centre for Data Science and AI, Indian Institute of Technology, Madras}
\city{Chennai}
\state{Tamil Nadu}
\country{India}
\postcode{600036}
}
\email{cs18d016@smail.iitm.ac.in}

\author{Akash Kumar Mohankumar}
\affiliation{
\institution{Indian Institute of Technology, Madras}
\city{Chennai}
\state{Tamil Nadu}
\country{India}
\postcode{600036}
}
\email{makashkumar99@gmail.com}

\author{Mitesh M. Khapra}
\affiliation{
\institution{Robert-Bosch Centre for Data Science and AI, Indian Institute of Technology, Madras}
\city{Chennai}
\state{Tamil Nadu}
\country{India}
\postcode{600036}
}
\email{miteshk@cse.iitm.ac.in}

\renewcommand{\shortauthors}{Sai, et al.}
\newcommand*{\mysqrt}[4]{\sqrt[\leftroot{#1}\uproot{#2}#3]{#4}}

\newcommand*{\MK}[1]{\textcolor{red}{}}
\newcommand*{\M}[1]{\textcolor{black}{#1}}
\newcommand*{\Ana}[1]{\textcolor{purple}{}} 
\newcommand*{\newA}[1]{\textcolor{black}{#1}} 
\newcommand*{\Akash}[1]{\textcolor{blue}{}}
\begin{abstract}
The success of Deep Learning has created a surge in interest in a wide range of Natural Language Generation (NLG) tasks. Deep Learning has not only pushed the state of the art in several existing NLG tasks but has also facilitated researchers to explore various newer NLG tasks such as image captioning. Such rapid progress in NLG has necessitated the development of accurate automatic evaluation metrics that would allow us to track the progress in the field of NLG. However, unlike classification tasks, automatically evaluating NLG systems in itself is a huge challenge. Several works have shown that early heuristic-based metrics such as BLEU, ROUGE are inadequate for capturing the nuances in the different NLG tasks. The expanding number of NLG models and the shortcomings of the current metrics has led to a rapid surge in the number of evaluation metrics proposed since 2014.  Moreover, various evaluation metrics have shifted from using pre-determined heuristic-based formulae to trained transformer models. This rapid change in a relatively short time has led to the need for a survey of the existing NLG metrics to help existing and new researchers to quickly come up to speed with the developments that have happened in NLG evaluation in the last few years. Through this survey, we first wish to highlight the challenges and difficulties in automatically evaluating NLG systems. Then, we  provide a coherent taxonomy of the evaluation metrics to organize the existing metrics and to better understand the developments in the field. We also describe the different metrics in detail and highlight their key contributions. Later, we discuss the main shortcomings identified in the existing metrics and describe the methodology used to evaluate evaluation metrics. Finally, we discuss our suggestions and recommendations on the next steps forward to improve the automatic evaluation metrics. 
\end{abstract}





\maketitle


\section{Introduction}
Natural Language Generation (NLG) refers to the process of automatically generating human-understandable text in one or more natural languages. The ability of a machine to generate such natural language text which is indistinguishable from that generated by humans is considered to be a pre-requisite for Artificial General Intelligence (AGI) - the holy grail of AI. Indeed, the Turing test \cite{10.1093/mind/LIX.236.433}, widely considered to be the ultimate test of a machine's ability to exhibit human-like intelligent behaviour requires a machine to have natural language conversations with a human evaluator. A machine would pass the test if the evaluator is unable to determine whether the responses are being generated by a human or a machine.
Several attempts have been made, but no machine has been able to convincingly pass the Turing test in the past 70 years since it was proposed. However, steady progress has been made in the field in the past 70 years with remarkable achievements in the past few years since the advent of Deep Learning \cite{goodfellow2016deeplearning_textbook, goldberg2016primer, young2018recent_trends_in_DL_based_NLP, deng2018deep_DL_in_NLP_book}. 

Indeed, we have come a long way since the early days of AI, when the interest in NLG was limited to developing rule based machine translation systems \cite{Hutchins97fromfirst} and dialog systems \cite{weizenbaum1966eliza, winograd1971procedures_SHRDLU, winograd1972understanding_natural_language}.
The earliest demonstration of the ability of a machine to translate sentences was the Georgetown-IBM Experiment where an IBM 701 mainframe computer was used to translate 60 Russian sentences into English \cite{Hutchins97fromfirst}. The computer used a rule based system with just six grammar rules and a vocabulary of 250 words. Compare this to the modern neural machine translation systems which get trained using millions of parallel sentences on multiple TPUs using a vocabulary of around 100K words \cite{DBLP:conf/nips/VaswaniSPUJGKP17_attention_is_all_you_need_transformers}. 
The transition to such mammoth data driven models is the result of two major revolutions that the field of Natural Language Processing (which includes Natural Language Understanding and  Natural Language Generation) has seen in the last five decades. The first being the introduction of machine learning based models in the late 1980s which led to the development of data driven models which derived insights from corpora. This trend continued with the introduction of Decision Trees, Support Vector Machines and statistical models like Hidden Markov Models, the IBM translation model, Maximum Entropy Markov Models, and Conditional Random Fields, which collectively dominated NLP research for at least two decades. The second major revolution was the introduction of deep neural network based models which were able to learn from large amounts of data and establish new state of the art results on a wide variety of tasks \cite{young2018recent_trends_in_DL_based_NLP, deng2018deep_DL_in_NLP_book}. 

The advent of Deep Learning has not only pushed the state of the art in existing NLG tasks but has created interest in solving newer tasks such as image captioning, video captioning, etc. Indeed, today NLG includes a much wider variety of tasks such as machine translation, automatic summarization, table-to-text generation (more formally, structured data to text generation), dialogue generation, free-form question answering, automatic question generation, image/video captioning, grammar correction, automatic code generation, \textit{etc}. This wider interest in NLG is aptly demonstrated by the latest GPT-3 model \cite{brown2020language_gpt3} which can write poems, oped-articles, stories and code (among other things). This success in NLP, in general, and NLG in particular, is largely due to 3 factors: (i) the development of datasets and benchmarks which allow training and evaluating models to track progress in the field (ii) the advancements in Deep Learning which have helped stabilise and accelerate the training of large models and (iii) the availability of powerful and relatively cheaper compute infrastructure on the cloud \footnote{GCP: https://cloud.google.com/ AWS: https://aws.amazon.com/ Azure: https://azure.microsoft.com/}. Of course, despite these developments, we are still far from developing a machine which can pass the Turing test or a machine which serves as the fictional Babel fish\footnote{Hitchhiker's Guide to the Galaxy} with the ability to accurately translate from one language to any other language. However, there is no doubt that we have made remarkable progress in the last seven decades. 

This brings us to the important question of ``tracking progress'' in the field of NLG. How does one convincingly argue that a new NLG system is indeed better than existing state-of-the-art systems? The ideal way of doing this is to show multiple outputs generated by such a system to humans and ask them to assign a score to the outputs. The scores could either be absolute or relative to existing systems. Such scores provided by multiple humans can then be appropriately aggregated to provide a ranking of the systems. However, this requires skilled annotators and elaborate guidelines which makes it a time consuming and expensive task. Such human evaluations can act as a severe bottleneck, preventing rapid progress in the field. For example, after every small change to the model, if researchers were to wait for a few days for the human evaluation results to come back, then this would act as a significant impediment to their work. Given this challenge, the community has settled for automatic evaluation metrics, such as BLEU \cite{DBLP:conf/acl/PapineniRWZ02_bleu}, which assign a score to the outputs generated by a system and provide a quick and easy means of comparing different systems and tracking progress. 

\begin{figure}[h]
    \centering
    \includegraphics[scale=0.27]{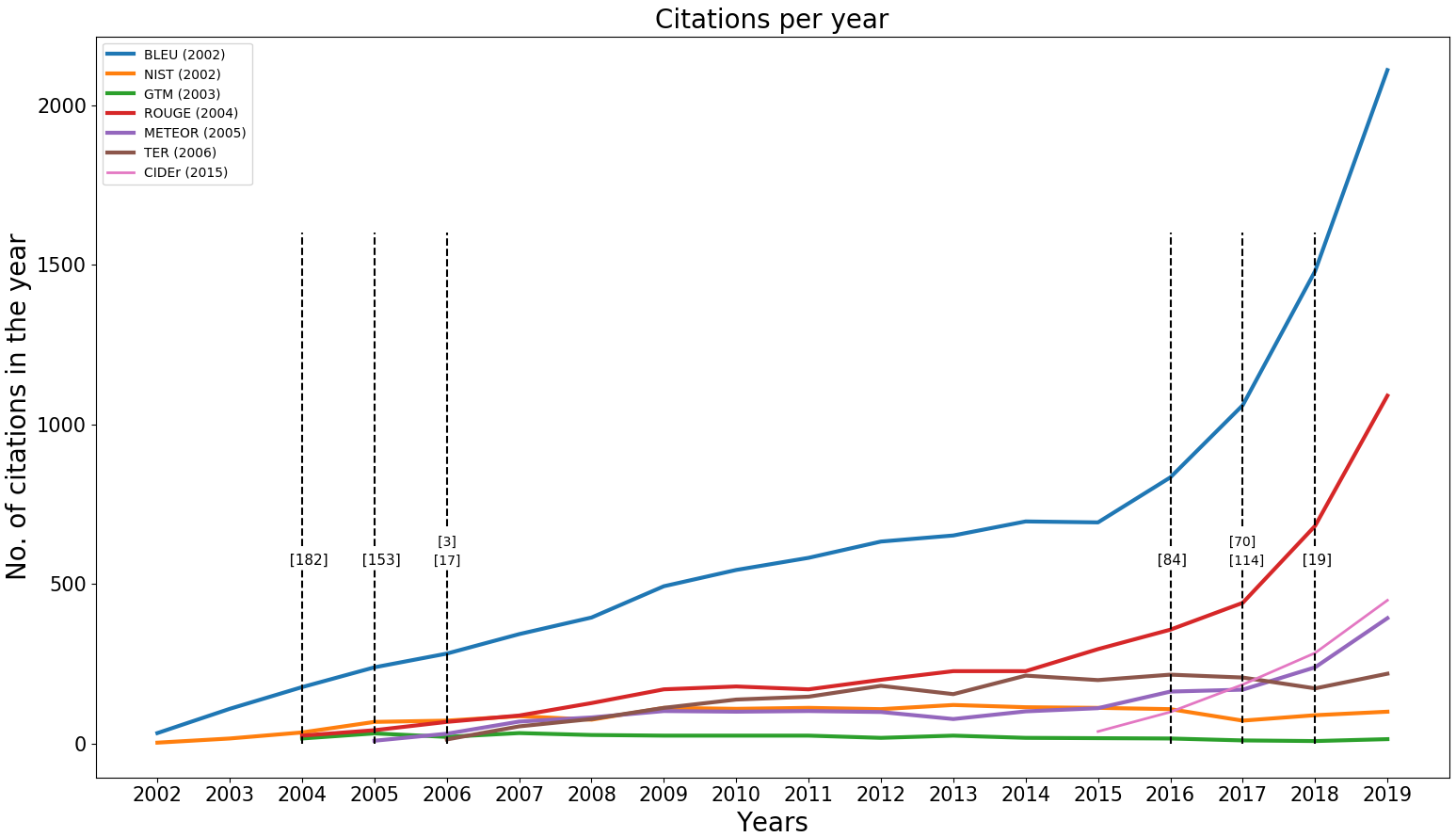}
    \caption{Number of citations per year on a few popular metrics. Dashed lines represent some of the major criticisms on these metrics at the corresponding year of publication.}
    \label{fig:init_metrics}
\end{figure}

Despite receiving their fair share of criticism, automatic metrics such as BLEU, METEOR, ROUGE, \textit{etc.}, continued to remain widely popular simply because there was no other feasible alternative. In particular, despite several studies \cite{DBLP:conf/lrec/ZhangVW04_bleu_nist_quant_imprvment,10.1007/978-3-540-30586-6_38_critique1,Ananthakrishnan2006SomeII_more_blues, callison-burch-etal-2006-evaluating_bleu_in_mt} showing that BLEU and similar metrics do not correlate well with human judgements, there was no decline in their popularity. This is illustrated in Figure \ref{fig:init_metrics} plotting the number of citations per year on some of the initial metrics from the time they were proposed up to recent years. The dashed lines indicate the years in which some of the major criticisms were published on these metrics, which, however, did not impact the adoption of these metrics.
\\
On the contrary as newer tasks like image captioning, question generation, dialogue generation became popular, these metrics were readily adopted for these tasks too. However, it soon became increasingly clear that such adoption is often not prudent given that these metrics were not designed for the newer tasks for which they are being adopted. For example, \citet{DBLP:conf/emnlp/NemaK18_qbleu} show that for the task of automatic question generation, it is important that the generated question is ``answerable'' and faithful to the entities present in the passage/sentence from which the question is being generated. Clearly, a metric like BLEU is not adequate for this task as it was not designed for checking ``answerability''. Similarly, in a goal oriented dialog system, it is important that the output is not only fluent but also leads to goal fulfillment (something which BLEU was not designed for). \Akash{Should we give some other example instead of goal oriented dialog system? Although BLEU might be used here, but I think researchers primarily rely on extrinsic evaluation methods to evaluate whether the task is fulfilled or not\MK{Yes, I agree. Please replace by an apt example.}}

Summarising the above discussion and looking back at the period from 2014-2016 we make 3 important observations (i) the success of Deep Learning had created an interest in a wider variety of NLG tasks (ii) it was still infeasible to do human evaluations at scale and (iii) existing automatic metrics were proving to be inadequate for capturing the nuances of a diverse set of tasks. This created a fertile ground for research in automatic evaluation metrics for NLG. Indeed, there has been a rapid surge in the number of evaluation metrics proposed since 2014. 
It is interesting to note that from 2002 (when BLEU was proposed) to 2014 (when Deep Learning became popular) there were only about 10 automatic NLG evaluation metrics in use. Since 2015, a total of atleast 36 new metrics have been proposed. In addition to earlier rule-based or heuristic based metrics such as Word Error Rate (WER), BLEU, METEOR and ROUGE, we now have metrics which exhibit one or more of the following characteristics: (i) use (contextualized) word embeddings \cite{greedymatch,vectorextrema,DBLP:conf/acl/MathurBC19_BERTr_contextualized_embeddings_translation,bertscore} (ii) are pre-trained on large amounts of unlabeled corpus (e.g. monolingual corpus in MT \cite{DBLP:journals/corr/abs-2004-04696_bleurt} or Reddit conversations in dialogue) 
(iii) are fine-tuned on task-specific annotated data containing human judgements \cite{DBLP:conf/acl/LoweNSABP17ADEM} and (iv) capture task specific nuances \cite{DBLP:conf/emnlp/NemaK18_qbleu, DBLP:conf/acl/DhingraFPCDC19_parent}. This rapid surge in a relatively short time has lead to the need for a survey of existing NLG metrics. Such a survey would help existing and new researchers to quickly come up to speed with the developments that have happened in the last few years. \MK{Somewhere in this paragraph add a sentence about the WMT shared task on evaluation. Also sneak in a sentence about the ``Evaluating evaluation metrics'' which received an honorable mention at ACL 2020 saying that this suggests that this area is really important.} 

\subsection{Goals of this survey}
The goals of this survey can be summarised as follows:

\begin{itemize}
    \item \textbf{Highlighting challenges in evaluating NLG systems:} The first goal of this work is to make the readers aware that evaluating NLG systems is indeed a challenging task. To do so, in section \ref{sec:nlg_tasks} we first introduce popular NLG tasks ranging from machine translation to image captioning. For each task, we provide examples containing an input coupled with correct and incorrect responses. Using these examples, we show that distinguishing between correct and incorrect responses is a nuanced task requiring knowledge about the language, the domain and the task at hand. Further, in section \ref{sec:human_eval} we provide a list of factors to be considered while evaluating NLG systems. For example, while evaluating an abstractive summarisation system one has to ensure that the generated summary is informative, non-redundant, coherent and have a good structure. The main objective of this section is to highlight that these criteria vary widely across different NLG tasks thereby ruling out the possibility of having a single metric which can be reused across multiple tasks.
    \item \textbf{Creating a taxonomy of existing metrics:} As mentioned earlier, the last few years have been very productive for this field with a large number of metrics being proposed. Given this situation, it is important to organise these different metrics in a coherent taxonomy based on the methodologies they use. For example, some of these metrics use the context (input) for judging the appropriateness of the generated output whereas others do not. Similarly, some of these metrics are supervised and require training data whereas others do not. The supervised metrics further differ in the features they use. We propose a taxonomy to not only organise existing metrics but also to better understand current and future developments in this field. We provide this taxonomy in section \ref{sec:taxonomy} and then further describe these metrics in detail in section \ref{sec:context_free_metrics} and \ref{sec:context_dependent_metrics}.
    \item \textbf{Understanding shortcomings of existing metrics:} While automatic evaluation metrics have been widely adopted, there have been several works which have criticised their use by pointing out their shortcomings. To make the reader aware of these shortcomings, we survey these works and summarise their main findings in section \ref{sec:criticism}. In particular, we highlight that existing NLG metrics have poor correlations with human judgements, are uninterpretable, have certain biases and fail to capture nuances in language.
    \item \textbf{Examining the measures used for evaluating evaluation metrics:}
    With the increasing number of proposed automatic evaluation metrics, it is important to assess how well these different metrics perform at evaluating NLG outputs and systems. We highlight the various methods used to assess the NLG metrics in section \ref{sec:evaluating_metrics}. We discuss the different correlations measures used to analyze the extent to which automatic evaluation metrics agree with human judgements. We then underscore the need to perform statistical hypothesis tests to validate the significance of these human evaluation studies. Finally, we also discuss some recent attempts to evaluate the adversarial robustness of the automatic evaluation metrics. 
    \item \textbf{Recommending next steps:} Lastly, we discuss our suggestions and recommendations to the community on the next steps forward towards improving automated evaluations. We emphasise the need to perform a more fine-grained evaluation based on the various criteria for a particular task. We highlight the fact that most of the existing metrics are not interpretable and emphasise the need to develop self-explainable evaluation metrics. We also point out that more datasets specific to automated evaluation, containing human judgements on various criteria, should be developed for better progress and reproducibility. 
\end{itemize}

\section{Various NLG Tasks}
\label{sec:nlg_tasks}
\begin{table}
\resizebox{1.0\textwidth}{!}{
\begin{tabular}{|l|p{10.5cm}|p{10.5cm}|}
\hline
\multicolumn{1}{|c|}{\textbf{Task}}   & \multicolumn{1}{c|}{\textbf{Input}} & \multicolumn{1}{c|}{\textbf{Example Generated Outputs}}                                                                                                                                                                                                                                                                                                                                                                                                                                                                                                                                                                                                                                                                                                                                                \\ \hline

\begin{tabular}[c]{@{}l@{}}\textbf{Machine Translation}\\ (French to English).\end{tabular}
& 
\begin{tabular}[c]{p{10cm}}
\textbf{French Source:} le pamplemousse est mon fruit le plus aimé mais la banane est son plus aimé.\\ \\ \textbf{English Reference:} The grapefruit is my most loved fruit  but the banana is her most loved.\end{tabular}                                                                                                                                                                                                                                                                                                                                                                                                                                                                                                                                                                                                                                                                                                                                                                                                                                    & \begin{tabular}[c]{p{10cm}}1. The grapefruit is my most expensive fruit but the banana is her most loved.\\ \\ 2. Grapefruit is my favorite fruit, but banana is her most beloved.\end{tabular}                                                                                                                                                                                                                                                                                                                                                                                                                                                                                                                                                                \\ \hline
\textbf{Abstractive Summarization}    & \begin{tabular}[c]{p{10cm}}\textbf{Document:} West Berkshire Council is setting up an emotional health academy to train psychology graduates and health professionals. The  local authority said, once trained, its staff will work with children, families, and schools. It wants to greatly reduce the wait mental health patients face from 12 months to less than a week. The council also hopes the new academy will stop problems escalating to the stage where they require attention from more highly trained mental health specialists. Director of Children 's Services Rachael Wardell said:  "It works better if you get in there sooner when people are waiting for help their condition gets worse. [...]\\ \\  \textbf{Reference Summary:} West Berkshire Council is setting up an emotional health academy to train psychology graduates and health professionals.\end{tabular} & \begin{tabular}[c]{p{10cm}}1. A mental health academy in Berkshire has been put up for sale in a bid to reduce the number of mental health patients. \\ \\ 2. West Berkshire Council aims to reduce the wait mental health patients face from 12 months to less than a week.\\ \\ 3. Plans to improve children's mental health services by setting up an emotional health academy in West Berkshire have been announced by the county's council.\end{tabular}                                                                                                                                                                                                                                                                                                                                                          \\ \hline
\textbf{Free-form Question Answering} & \begin{tabular}[c]{p{10cm}}\textbf{Question:} How do Jellyfish function without brains or nervous systems? {[}...{]}\\ \\ \textbf{Documents:}  {[}...{]} Jellyfish do not have brains, and most barely have nervous systems. They have primitive nerve cells that help them orient  themselves in the water and sense light and touch. {[}...{]} While they don't possess brains, the animals still have neurons that send all sorts of signals throughout their body. {[}...{]} They may accomplish this through the assistance of their nerve rings. Jellyfish don’t have brains, and that’s just where things begin. They don’t have many of the body parts that are typical in other animals. {[}...{]} \end{tabular}                                                                                                                                                                                                                                                                                                                            & \begin{tabular}[c]{p{10cm}} 1. Jellyfish do actually have a brain though they are small relative to their body size. They can also generate and process sensory information using their own internal network of neurons. \\ \\

2. Jellyfish is a very simple creature. They’re invertebrate: creatures without a backbone. Most jellyfish have really short life spans. Sometimes just a couple of hours. \\ \\ 
3. Jellyfish may not have a brain, but they have a rough nervous system and innate behaviors. They use their nerve cells to detect light, chemicals and movements. They detect stimuli, and transmits impulses both throughout the nerve net and around a circular nerve ring, to other nerve cells.
.\end{tabular} \\ \hline

\multirow{3}{*}[-13em]{\textbf{Question Generation}} & \begin{tabular}[c]{p{10cm}}
\textbf{Reading Comprehension Question Generation}\\
\textbf{Passage:} Liberated by Napoleon’s army in 1806, Warsaw was made the capital of the newly created Duchy of Warsaw\\  \textbf{Answer:} Napoleon’s
\end{tabular}                                                                                                           
& \begin{tabular}[c]{p{10cm}}
1. What was the capital of the newly duchy of Warsaw?\\
2. When was warsaw liberated by Napoleon’s army. \\
3. Who liberated Warsaw in 1806?\\
4. Whose army liberated Warsaw in 1806?
\end{tabular}  \\ \cline{2-3}
 & \begin{tabular}[c]{p{10cm}}
\textbf{Visual Question Generation}\\
\textbf{Image:}\\ \includegraphics[width=7cm]{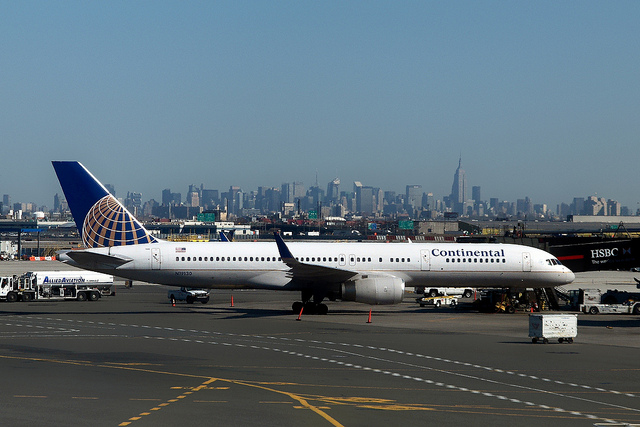}\ \\  \textbf{Answer:} Airport
\end{tabular}                                                                                                           
& \begin{tabular}[c]{p{10cm}}
1. Where is the motorbike located?\\
2. What is the color of the bike?\\
3. What is the color of water below the aeroplane?\\
4. What is located on the runway? \\
5. What time of day is it? \\
6. What is the scene located? \\
7. Where is the scene located? \\
8. Where is the aeroplane situated?
\end{tabular}  \\ \cline{2-3}
  & \begin{tabular}[c]{p{10cm}}
\textbf{Knowledge Base Question Generation}\\
\textbf{KB Entries:}\\
Blade Runner \textit{directed\_by} Ridley Scott\\
Blade Runner \textit{written\_by} Philip K. Dick, Hampton Fancher\\
Blade Runner \textit{starred\_actors} Harrison Ford, Sean Young, $\dots$ \\
Blade Runner \textit{release\_year} 1982\\
Blade Runner \textit{has\_tags} dystopian, noir, police, androids $\dots$
\end{tabular}                                                                                                           
& \begin{tabular}[c]{p{10cm}}
1. What role does Harrison Ford play in the movie Blade Runner?\\
2. What is the plot of the film Blade Runner?\\
3. How was the reception to the movie Blade Runner?\\
4. What year was the movie Blade Runner released?\\
5. Who is the writer of the film Blade Runner?\\
6. Can you describe the movie Blade Runner in a few words?
\end{tabular}  \\ \hline

\textbf{Data to Text Generation}      &                                                                                  \begin{tabular}[c]{p{10cm}}
\textbf{Data:} \\
(John E Blaha \textit{birthdate} 1942 08 26)\\
(John E Blaha \textit{birthplace} San Antonio)\\
(John E Blaha \textit{occupation} Fighter Pilot) \\

\textbf{Reference Text:} John E Blaha, born in San Antonio on 1942-08-26, worked as a fighter pilot
\end{tabular}                             & \begin{tabular}[c]{p{10cm}}1. John E Blaha who worked as a fighter pilot was born on 26.08.1942.
\\ 2. Fighter pilot John E Blaha was born in San Antonio on the 26th July 1942. \\
3. John E Blaha, born on the 26th of August 1942 in San Antonio, served as a fighter pilot.
\end{tabular}                                           \\ \hline
\textbf{Dialogue Generation} & \begin{tabular}[c]{p{10cm}}
\textbf{Context:}\\
First Speaker: Can you do push-ups?\\ Second Speaker: Of course I can. It’s a piece of cake! Believe it or not, I can do 30 push-ups a minute.\\ First Speaker: Really? I think that’s impossible!\\ Second Speaker: You mean 30 push-ups?\\ First Speaker: Yeah!\end{tabular}                                                                                                                                                                                                                                                                                                                                                                                                                                                                                                                                                                                                                                                                                                                                                                                                                          & \begin{tabular}[c]{p{10cm}}
1. Second Speaker: Would you like to eat a piece of cake before gym?\\
2. Second Speaker: Of course I can. It’s a piece of cake! Believe it or not, I can do 30 push-ups a minute.\\
3. Second Speaker: Hmm.. okay.\\
4.  Second Speaker: Start your timer, here we go.\\
5. Second Speaker: You don’t know that I am a fitness trainer, do you?\\
6. Second Speaker: Haha, you are right, was just kidding!
\end{tabular}                                                                                                                                                                                                                                                                                                                                                                                                                                                                                                                                                    \\ \hline
\textbf{Image captioning} &  
\begin{tabular}[c]{p{10cm}}\textbf{Image:} \\ \includegraphics[width=7cm]{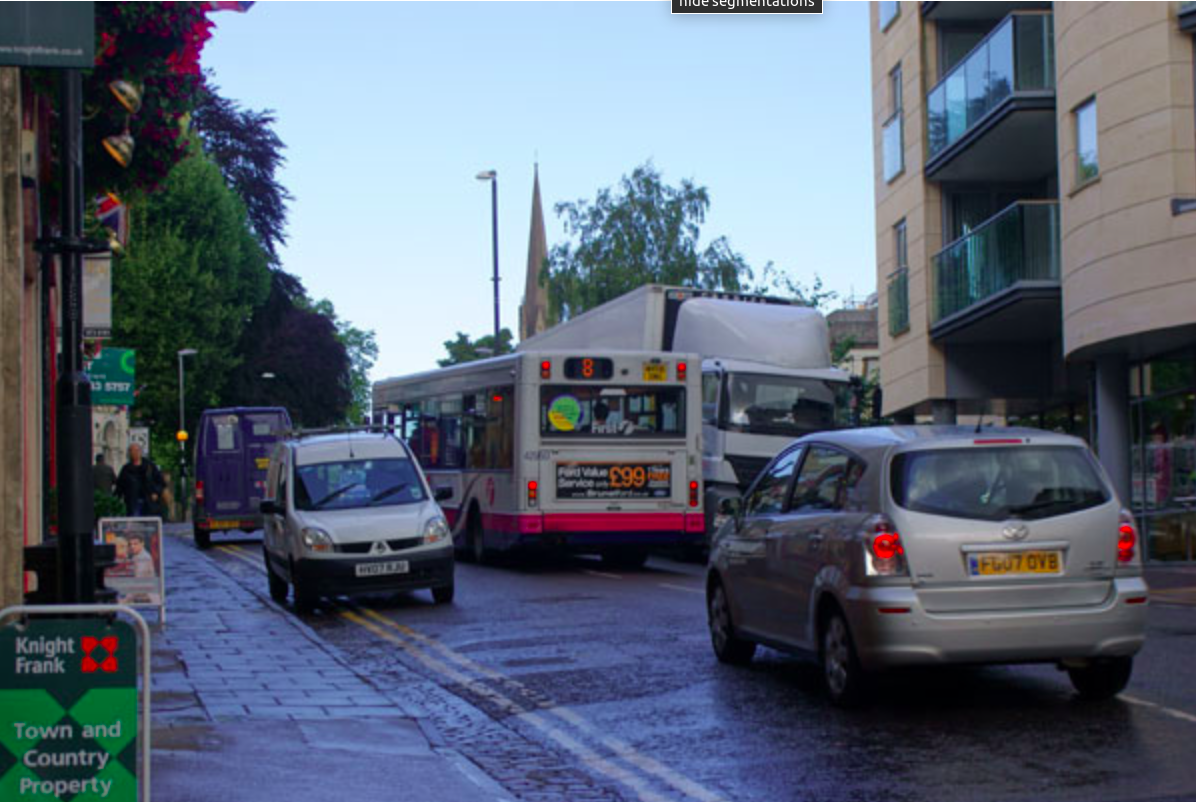}\\ \\ \textbf{Reference Caption:} Bus, truck and cars going down a city street.\end{tabular}
&
\begin{tabular}[c]{p{10cm}}
1. People are walking under umbrellas on a city street. \\
2. A cloudy sky over a city street \\
3. The cars and trucks are headed down the street with a view of the scenic valley and mountain range.\\
4. A white bus sits on the road in a street. \\
5. A long bus is going down the street. \\
6. A street is shown with a car travelling down it.\\
7. A city bus traveling down the street next to a truck and car.\\
8. A crowded city street where cars, bus and truck are facing both directions in the same lane. \\
9. Cars, truck and bus moving on a road with green trees and buildings on the side. \\
10. Two grey cars travelling opposite to each other along with a white bus and grey truck on a road with buildings, and trees. 
\end{tabular}
\\ \hline
\end{tabular}
}
\caption{Examples inputs and generated outputs for various Natural Language Generation tasks.}
\label{tab:examples_nlg}
\end{table}
In this section, we describe various NLG tasks and highlight the challenges in automatically evaluating them with the help of examples in Table \ref{tab:examples_nlg}. We shall keep the discussion in this section slightly informal and rely on examples to build an intuition for why it is challenging to evaluate NLG systems. Later on, in section \ref{sec:human_eval}, for each NLG task discussed below, we will formally list down the criteria used by humans for evaluating NLG systems. We hope that these two sections would collectively reinforce the idea that evaluating NLG systems is indeed challenging since the generated output is required to satisfy a wide variety of criteria across different tasks. \\
\\
\textbf{Machine Translation (MT)} refers to the task of converting a sentence/document from a source language to a target language. The target text should be fluent, and should contain all the information in the source text without introducing any additional details. The challenge here is that there may be many alternative correct translations for a single source text and usually only a few gold standard reference translations are available. Further, translations with a higher word-overlap with the gold standard reference need not have a better translation quality. For example, consider the two translations shown in the first row of Table \ref{tab:examples_nlg}. Although translation 1 is the same as the reference except for one word, it does not express the same meaning as the reference/source. On the other hand, translation 2 with a lower word overlap has much better translation quality. A good evaluation metric should thus be able to understand that even changing a few words can completely alter the meaning of a sentence. Further, it should also be aware that certain word/phrase substitutions are allowed in certain situations but not in others. For example, it is perfectly fine to replace ``loved'' by ``favorite'' in the above example but it would be inappropriate to do so in the sentence ``I loved him''. Of course, in addition, a good evaluation metric should also be able to check for the grammatical correctness of the generated sentence (this is required for all the NLG tasks listed below).
\\\\
\textbf{Abstractive Summarization (AS)} is the task of shortening a source document to create a summary using novel phrases that concisely represent the contents of the source document. The summary should be fluent, consistent with the source document, and concisely represent the most important/relevant information within the source document. In comparison to MT, there can be much greater diversity between valid outputs (summaries) for a given input (source document), and hence evaluation can be even more difficult. Further, unlike MT, the summary need not contain all the information present in the source document. However, it has to be coherent and must highlight the important information in the source document. For example, consider the source document and summaries in Table \ref{tab:examples_nlg}. Summary 1 is not consistent with the source document (\textit{i.e.}, is factually incorrect) though it contains important words and entities present in the source document. While summary 2 is consistent with the provided document, it does not convey the crucial information that the council is going to set up a health academy. On the other hand, summary 3 is of much better quality though it is phrased very differently from the provided reference. A good evaluation metric should thus be able to distinguish between (i) summaries which have a good word overlap with the source document and/or reference summary but are factually incorrect, (ii) summaries which are factually correct but missing crucial information, and (iii) summaries which are factually correct and contain adequate information even when they are worded differently from the reference summary. 
\\\\
\textbf{Free-form Question Answering (QA)} refers to the task of generating an answer in natural language, as opposed to selecting a span within a text to answer a given question. The task may additionally include background information in the form of a document, knowledge base, or image. Like the previously discussed tasks, the answer to a given question can be phrased in different ways. The evaluation metric should identify whether the answer is fluent, addresses the given question and is consistent with the provided background information or not. For example, the first answer in Table \ref{tab:examples_nlg} addresses the given question, but it is factually incorrect and inconsistent with the provided document. While the second answer is factually correct, it does not address the specific question. The last answer both addresses the question and is consistent with the provided passage.  
\\\\
\textbf{Question Generation (QG)} refers to the task of crafting a question based on an input source and optionally an answer. The input source could be a document, a knowledge base, or an image. The generated question should be fluent, answerable from the input source, and specific to the answer (if provided). Consider the reading comprehension based question generation example in Table \ref{tab:examples_nlg}; question 1 is grammatically incorrect and not specific to the given answer. Question 2 is fluent but not specific to the answer, whereas question 3, 4 are fluent, answerable from the passage, and specific to the given answer. The main challenge here is that a good evaluation metric should be able to identify whether the generated question adheres to \textit{all} these varied requirements or not. Further, evaluation can be more challenging when the task requires multi-modal understanding. For instance, in the visual question generation example in Table \ref{tab:examples_nlg}, the evaluation metric has to identify that questions 1, 2, and 3 cannot be answered from the provided image. Similarly, questions 4 and 5 are not specific to the given answer, and question 6 is not fluent. Questions 7 and 8 are both appropriate questions for the given example. In some settings, the answer may not be provided as an input, as illustrated in the example in Table \ref{tab:examples_nlg} where a question needs to be generated from a knowledge base. In this example, questions 1, 2, and 3 are not answerable from the provided knowledge base even though the entities contained in these questions are present in the knowledge base. Questions 4, 5, and 6, on the other hand, are appropriate questions for the given knowledge base (i.e, they are all fluent and answerable from the input source). Note that to assign a high score to Question 6, the evaluation metric should also have some domain/common sense knowledge to understand that ``tags'' correspond to ``short descriptions''.
\\\\
\textbf{Data to Text Generation (D2T)} refers to the task of producing natural language text from a structured or semi-structured data source. The data source can either be a database of records, a spreadsheet, a knowledge graph, \textit{etc}. In this task, a good evaluation metric is required to judge that the generated text is fluent, adequately verbalized, factually correct and covers all relevant facts in the data source. Consider the example in Table \ref{tab:examples_nlg}. The first sentence does not cover all the facts mentioned in the provided data source (birthplace is missing). The second sentence is factually incorrect (birth date is incorrectly verbalized). The third sentence is an appropriate description as it is fluent and accurately covers all fields in the data source. Even though it is worded differently when compared to the given reference, a good evaluation metric should not penalise it for this alternative phrasing.
\\\\
\textbf{Dialogue Generation (DG)} refers to the task of having conversations with human beings. The conversations could be open-ended or targeted to accomplish some specific goals. Each generated response should be fluent, coherent with the previous utterances, and aligned with the specific goal (if any). Additionally, it is also desired that the dialogue agent makes the conversation interesting and engaging while also displaying a consistent persona. In the example open-domain conversation mentioned in Table \ref{tab:examples_nlg}, the first response is not coherent with the context although it contains words and phrases which are present in the context (``piece of cake'', ``gym''). The second response, although being coherent with the context, is an exact repetition of one of the already generated responses and hence makes the conversation monotonous (not interesting/engaging). The third response is very short and vague, and therefore would again result in a boring conversation. The last three responses can be considered as valid responses to the given context. Note that the last three responses are very diverse, carrying different meanings but can still be considered appropriate responses to the conversation. Indeed, the biggest challenge in evaluating dialogue generation systems is that an evaluation metric should allow for multiple varied responses for the same context. Further, it should also judge other parameters such as fluency, coherence, interestingness, consistency (in persona), \textit{etc}.
\\\\
\textbf{Image Captioning (IC)} is the task of generating a textual description of a given image. The generated caption must be fluent and adequately represent the important information in the image. Consider the example in Table \ref{tab:examples_nlg}. The first and second captions are clearly not consistent with the given image. The third caption is partially consistent; the details of the valley and mountain are not consistent with the image.  Captions 4, 5, and 6 are consistent with the image, but they are incomplete. They do not describe the presence of other vehicles in the image. The captions 7 to 10 appropriately describe the important information in the given image. As we can observe, it is possible to have concise captions like 7 or very descriptive captions like 10. It is not necessary that the caption should cover all the elements in the image. For example, it is perfectly fine for a caption to ignore objects in the background like (sky, grass, etc) and still provide a meaningful description of the image. Thus a good evaluation metric must check that the generated caption is fluent, contains the important entities in the image, and accurately describes the relation between them (e.g., ``boy throwing a ball'' v/s ``boy catching a ball''). Further, it should not be biased towards longer captions which may contain unnecessary details (e.g., ``sky in the background'') and should be fair to shorter captions which concisely and accurately describe the image. \\

Apart from the tasks mentioned above, there are several other NLG tasks such as spelling and grammar correction, automatic paraphrase generation, video captioning, simplification of complex texts, automatic code generation, humour generation, \textit{etc} \cite{DBLP:journals/jair/GattK18_survey_sota_nlg}. However, we limit the above discussion to the most popular and well-studied tasks as most evaluation metrics have been proposed/studied in the context of these tasks.



\section{Human evaluation of NLG Systems}
\label{sec:human_eval}
As mentioned earlier, the ideal way of evaluating an NLG system is to ask humans to evaluate the outputs generated by the system. In this section, we first describe the procedure used for such an evaluation. Next, we supplement the anecdotal discussion in the previous section, by listing down and concretely defining the desired qualities in the output for different NLG tasks. By doing so, we hope to convince the readers that evaluating NLG systems is a multi-faceted task requiring simultaneous assessment of a wide set of qualities.  
\subsection{Human Evaluation Setup}
Depending on the budget, availability of annotators, speed and required precision, different setups have been tried for evaluating NLG systems. The different factors to consider in such an evaluation setup are as follows:
\begin{itemize}
    \item \textbf{Type of evaluators:} The evaluators could be experts \cite{belz-reiter-2006-comparing_auto_human_nlg_eval}, crowdsourced annotators \cite{DBLP:conf/emnlp/Callison-Burch09_amt_fast_cheap,kryscinski-etal-2019-neural_summ_critic,DBLP:journals/corr/abs-1910-08684_wikibio_confidence}, or even end-users \cite{DBLP:conf/nips/GhandehariounSJ19_selfplay,DBLP:conf/naacl/SeeRKW19} depending on the requirements of the task and the goal of the evaluation. For example, for evaluating a translation system one could hire bilingual experts (expensive) or even monolingual experts (relatively less expensive). The monolingual experts could just compare the output to an available reference output whereas with bilingual experts such a reference output is not needed. Further, a bilingual expert will be able to better evaluate where the nuances in the source language are accurately captured in the target language. If the speed of evaluation is the primary concern then crowd-sourced workers can also be used. In such a situation, one has to be careful to provide very clear guidelines, vet the workers based on their past records, immediately weed out incompetent workers and have an additional layer of quality check (preferably with the help of 1-2 expert in-house annotators). Clearly such crowdsourced workers are not preferred in situations requiring domain knowledge - e.g., evaluating an NLG system which summarises financial documents. For certain tasks, such as dialogue generation, it is best to allow end-users to evaluate the system by engaging in a conversation with it. They are better suited to judge the real-world effectiveness of the system. 
    \item \textbf{Scale of evaluation:} The annotators are typically asked to rate the output on a fixed scale, with each number corresponding to a specific level of quality, called the Likert scale \cite{likert1932technique_likert_very_first_in_psychology}. In a typical Likert scale the numbers 1 to 5 would correspond to Very Poor, Poor, Okay, Good and Very Good. However, some works \cite{DBLP:conf/acllaw/GrahamBMZ13_DA_scores,DBLP:conf/eacl/GattB10_tuna,DBLP:conf/acl/BelzK11_continuous_eval} have also experimented with a dynamic/movable continuous scale that can allow the evaluator to give more nuanced judgements. 
    An alternate setting asks humans to assign a rating to the output based on the amount of post-editing required, if any, to make the output acceptable \cite{bojar-etal-2014-findings,DBLP:journals/corr/abs-1807-02202_chaganty}. The evaluators could also be asked for binary judgements rather than a rating to indicate whether a particular criteria is satisfied or not. This binary scale is sometimes preferred over a rating scale, which usually contains 5 or 7 rating points, in order to force judges to make a clear decision rather than give an average rating (by choosing a score at the middle of the scale) \cite{horbach-etal-2020-linguistic_approp}. By extension, any even-point rating scale could be used to avoid such indecisiveness.
    \item \textbf{Providing a reference and a context:} In many situations, in addition to providing the output generated by the system, it is helpful to also provide the context (input) and a set of reference outputs (if available). However, certain evaluations can be performed even without looking at the context or the reference output. For instance, evaluating fluency (grammatical correctness) of the generated sentence does not require a reference output. References are helpful when the evaluation criteria can be reduced to a problem of comparing the similarity of information contained in the two texts. For example, in most cases, a generated translation can be evaluated for soundness (coherence) and completeness (adequacy) by comparing with the reference (without even looking at the context). However, for most NLG tasks, a single reference is often not enough and the evaluator may benefit from looking at the context. The contexts contains much more information which is difficult to be captured by a small set of references. In particular, referring to the examples provided for ``Abstractive Summarisation'', ``Image Captioning'' and ``Dialogue Generation'' in Table \ref{tab:examples_nlg}, it is clear that it is difficult for the evaluator to do an accurate assessment by only looking at the generated output and the providing references. Of course, reading the context adds to the cognitive load of the evaluator but is often unavoidable. 
    \item \textbf{Absolute v/s relative evaluation :} The candidate output could be evaluated individually or by comparing it with other outputs. In an individual output evaluation, the candidate is provided an absolute rating for each desired criteria. On the other hand, in a comparison setup, an annotator could either be asked to simultaneously rate the multiple outputs (from competing systems) \cite{DBLP:conf/emnlp/NovikovaDCR17} or be asked to preferentially rank the multiple outputs presented \cite{DBLP:conf/cvpr/VedantamZP15_CIDEr,DBLP:conf/eacl/KilickayaEIE17_wmd_ic,DBLP:journals/csl/DusekNR20_e2enlg}. 
    This could also just be a pairwise comparison \cite{DBLP:journals/corr/abs-1909-03087_acute_eval,li2017adversarial,DBLP:conf/acl/DhingraFPCDC19_parent} of two systems. In such a setup, 
    the two systems are compared based on the number of times their outputs were preferred (wins), not preferred (losses), and equally preferred (ties).
    \item \textbf{Providing Rationale :} The evaluators might additionally be asked to provide reasons for their decisions, usually by highlighting the corresponding text that influenced the rating \cite{DBLP:journals/corr/abs-1807-02202_chaganty}. Such fine-grained feedback can often help in further improving the system. 
\end{itemize}
Irrespective of the setup being used, typically multiple evaluators are shown the same output and their scores are then aggregated to come up with a final score for each output or the whole system. The aggregate can be computed as a simple average or a weighted average wherein each annotator is weighted based on his/her past performance or agreement with other annotators \cite{10.5555/1756006.1859894}. In general, it is desired to have a high inter-annotator agreement (IAA), which is usually measured using Cohen's Kappa or Fleiss Kappa co-efficient or Krippendorff’s alpha.  
Alternatively, although not popularly, IAA could be measured using Jaccard similarity, or an F1-measure (based on precision and recall between annotators) \cite{bosch_2020}. 
Achieving a high-enough IAA is more difficult on some NLG tasks which have room for subjectivity \cite{amidei-etal-2018-rethinking_agreement_iaa}. A lower IAA can occur due to (i) human-error (ii) inadequacy of the guidelines or setup (iii) ambiguity in the text \cite{DBLP:journals/nle/SampsonB08}.  
To enhance IAA, \citet{DBLP:journals/corr/abs-1807-02202_chaganty} find that asking the evaluators to highlight the portion of the text that lead to their decision or rating helps in getting better agreement. Alternatively, \citet{DBLP:conf/emnlp/NemaK18_qbleu} arrange for a discussion between the annotators after the first round of evaluation, so as to mutually agree upon the criteria for the ratings.
To get a better IAA and hence a reliable evaluation, it is important that the human evaluators be provided with clear and sufficient guidelines. These guidelines vary across different NLG tasks as the criteria used for evaluation vary across different tasks, as explained in the next subsection.

\subsection{Criteria used for Evaluating NLG systems }
Most human evaluations are based on checking for task fulfillment, \textit{i.e.}, humans are asked to rate or compare the generated sentences (and the generating systems) to indicate how satisfactorily they meet the task requirements overall. However, evaluations can also be performed at a more fine-grained level where the various contributing factors are individually evaluated, \textit{i.e.}, the generated text is assigned a separate rating or ranking based on each of the desired qualities, independent of the other qualities/criteria. 
One such desired criteria is that the generated texts should have good `fluency'.
\textbf{Fluency} refers to correctness of the generated text with respect to grammar and word choice, including spellings. 
To check for fluency in the generated output, the evaluators might be asked the question, ``How do you judge the fluency of this text?" followed by a 5-point rating scale \cite{callison-burch-etal-2006-evaluating_bleu_in_mt}:
1. Incomprehensible 2. Not fluent German 3. Non-native German 4. Good German 5. Flawless German.
Instead of a 5-point scale, other scales with different quality ratings could be used: ``How natural is the English of the given sentence?'' 1. Very unnatural 2. Mostly unnatural 3. Mostly natural 4. Very natural \cite{DBLP:conf/naacl/SeeRKW19}. 
Another possibility is to present multiple candidate sentences and ask the evaluator, ``Which of these sentences seems more fluent?". The evaluator then indicates a preference ordering with ties allowed. 

Fluency in the generated output is a desired criteria for all the NLG tasks. However, the comprehensive list of criteria used for evaluation varies across different tasks. Hence, we discuss the set of criteria for each task separately now. Note that we have already defined fluency and mentioned that it is important for all NLG tasks. Hence, we do not discuss it again for each task independently. Further, note that the set of criteria is not standardized and some works use slightly different criteria/ sub-categorizations for the same task. Often the difference is only in the label/term used for the criteria but the spirit of the evaluation remains the same. Thus, for the below discussion, we consider only the most prominently used criteria for each task. In the discussion below and the rest of the paper, we interchangeably refer to the output of an NLG system as the hypothesis. 
\\\\
\textbf{Machine Translation:} Here, bilingual experts are presented with the source sentence and the hypothesis. Alternatively, monolingual experts can be presented with the reference sentence and the hypothesis. For each output, they are usually asked to check two important criteria: fluency and adequacy of the hypothesis \cite{DBLP:journals/nle/GrahamBMZ17_crowd_alone} as described below. 
\begin{itemize}
    \item \textbf{Adequacy:} The generated hypothesis should adequately represent all the information present in the reference. To judge adequacy a human evaluator can be asked the following question \cite{callison-burch-etal-2006-evaluating_bleu_in_mt}: How much of the meaning expressed in the reference translation is also expressed in the hypothesis translation? 1. None 2. Little 3. Much 4. Most 5. All
\end{itemize}
~\\
\textbf{Abstractive Summarization:} Human evaluators are shown the candidate summary along with the source document and/or a set of references. The evaluators are typically asked to rate informativeness and coherence \cite{DBLP:conf/ntcir/Mani01,mani-etal-1999-improving}. 
Alternatively, in a more elaborate evaluation the evaluators are asked to check for fluency, informativeness, non-redundancy, referential clarity, and structure \& coherence \cite{DBLP:journals/cai/SteinbergerJ09,DBLP:journals/lre/LloretPA18} as described below. 
\begin{itemize}
    \item \textbf{Informativeness:} The summary should convey the key points of the text. 
    For instance, a summary of a biography should contain the significant events of a person's life. We do not want a summary that only quotes the person's profession, nor do we want a summary that is unnecessarily long/verbose.
    \item \textbf{Non-redundancy:} The summary should not repeat any points, and ideally have maximal information coverage within the limited text length.
    \item \textbf{Referential clarity:} Any \newA{intra-sentence or cross-sentence} references in the summary should be unambiguous and within the scope of the summary. \newA{For example, if a pronoun is being used, the corresponding noun it refers to should also be present at some point before it in the summary. Also there should not be any ambiguities regarding the exact entity or information (such as a previous point) that is being referred to.} 
    \item \textbf{Focus:} The summary needs to have a focus and all the sentences need to contain information related to this focal point. For example, while summarising a news item about a Presidential debate, the focus of the summary could be the comments made by a candidate during the debate. If so, it should not contain irrelevant sentences about the venue of the debate.
    \item \textbf{Structure and Coherence:} The summary should be a well-organized and coherent body of information, not just a dump of related information. Specifically, the sentences should be connected to one another, maintaining good information flow.
\end{itemize}
~\\
\textbf{Question Answering:} Here, human evaluators are first presented with the question and the candidate answer to check if the answer is plausible \cite{DBLP:journals/corr/abs-1807-02202_chaganty}. Subsequently, the context passage/image is provided to check whether the answer is correct and consistent with the context. Alternatively, since question answering datasets are usually provided with gold standard answers for each question, the judges might simply be asked to report how closely the candidate answer captures the same information as the gold standard answer. The important criteria used for QA are fluency and correctness. 
\begin{itemize}
    \item \textbf{Correctness}: The answer should correctly address the question and be consistent with the contents of the source/context provided.
\end{itemize}
~\\
\textbf{Question Generation:} Here, the candidate questions are presented to the evaluators along with the context (passage/image, etc.) from which the questions were generated. This may be accompanied with a set of candidate answers \cite{DBLP:conf/naacl/Hosking019}, although if they are not provided, even when available in the dataset, it is to avoid creating any bias in the evaluator's mind \cite{DBLP:conf/emnlp/NemaK18_qbleu}. The evaluators are then asked to consider the following criteria\cite{DBLP:conf/emnlp/NemaK18_qbleu,DBLP:conf/naacl/Hosking019}:
\begin{itemize}
    \item \textbf{Answerability}: This is to determine whether the generated question is answerable given the context. A question might be deemed unanswerable due to its lack of completeness or sensibility, or even if the information required to answer the question is not found in the context. The latter could be acceptable in some scenarios where ``insufficient information'' is a legitimate answer (for example, if the questions are used in a quiz to check if the participants are able to recognize a case of insufficient information). However, generating too many such questions is undesirable and the evaluators may be asked to report if that is the case. 
    \item \textbf{Relevance}: This is to check if questions are related to the source material they are based upon. Questions that are highly relevant to the context are favoured. For example, a question based on common-sense or universal-facts might be answerable, but if it has no connection to the source material then it is not desired.
    \end{itemize}
~\\    
\textbf{Data to Text generation:} 
Here, human judges are shown the generated text along with the data (\textit{i.e.}, table, graph, etc). The criteria considered during human evaluation vary slightly in different works, such as WebNLG challenge \cite{shimorina2019webnlg_human_eval}, E2E NLG dataset \cite{DBLP:journals/csl/DusekNR20_e2enlg} or WikiBio dataset \cite{DBLP:journals/corr/abs-1910-08684_wikibio_confidence}.  
Here, we discuss the more fine-grained criteria of ``faithfulness'' and ``coverage'' as used in \cite{DBLP:journals/corr/abs-1910-08684_wikibio_confidence,DBLP:conf/acl/DhingraFPCDC19_parent} as opposed to the single criteria of ``semantic adequacy'' as used in \cite{shimorina2019webnlg_human_eval}.
\begin{itemize}
    \item \textbf{Faithfulness:} It is important for the text to preserve the facts represented in the data. For example, any text that misrepresents the year of birth of a person would be unacceptable and would also be ranked lower than a text that does not mention the year at all.
    \item \textbf{Informativeness or Coverage:} The text needs to adequately verbalize the information present in the data. As per the task requirements, coverage of all the details or the most significant details would be desired.
\end{itemize}
~\\
\textbf{Automated Dialogue:} For evaluating dialogue systems, humans are typically asked to consider a much broader set of criteria. One such exhaustive set of criteria as adopted by \cite{DBLP:conf/naacl/SeeRKW19}, is presented below along with the corresponding questions provided to the human evaluators:
\begin{itemize}
    \item \textbf{Making sense:} Does the bot say things that don't make sense?
    \item \textbf{Engagingness:} Is the dialogue agent enjoyable to talk to?
    \item \textbf{Interestingness:} Did you find the bot interesting to talk to?
    \item \textbf{Inquisitivenes:} Does the bot ask a good amount of questions?
    \item \textbf{Listening:} Does the bot pay attention to what you say?
    \item \textbf{Avoiding Repetition:} Does the bot repeat itself? (either within or across utterances)
    \item \textbf{Humanness:} Is the conversation with a person or a bot?
\end{itemize}
Often for dialogue evaluation, instead of separately evaluating all these factors, the evaluators are asked to simply rate the overall quality of the response \cite{DBLP:conf/acl/LoweNSABP17ADEM,DBLP:conf/aaai/TaoMZY18ruber}, or specifically asked to check for relevance of the response \cite{ghazarian19}. For task-oriented dialogues, additional constraints are taken into consideration, such as providing the appropriate information or service, guiding the conversation towards a desired end-goal, \textit{etc}. In open-domain dialogue settings also, additional constraints such as persona adherence \cite{personachat}, emotion-consistency \cite{DBLP:conf/nips/GhandehariounSJ19_selfplay}, \textit{etc}, are being used to expand the expectations and challenge the state-of-the-art.
~\\\\
\textbf{Image Captioning:} The captions are presented to the evaluators along with the corresponding images to check for relevance and thoroughness \cite{DBLP:journals/cviu/AdityaYBAF18_image_understanding}.
\begin{itemize}
    \item \textbf{Relevance:} This measures how well the caption is connected to the contents of the image. More relevance corresponds to a less-generic/more-specific caption that accurately describes the image. For example, the caption ``A sunny day'' is a very generic caption and can be applicable for a wide variety of images.
    \item \textbf{Thoroughness:} The caption needs to adequately describe the image. Usually the task does not require a complete description of everything in the image but the caption must cover the main subjects/actions in the image and not miss out any significant details.
\end{itemize}
~\\

In summary, the main takeaway from the above section is that evaluating NLG systems is a very nuanced task requiring multiple skilled evaluators and accurate guidelines which clearly outline the criteria to be used for evaluation. Further, the evaluation is typically much more than assigning a single score to the system or the generated output. In particular, it requires simultaneous assessment of multiple desired qualities in the output.

\section{Taxonomy of Automated Evaluation Metrics}
\label{sec:taxonomy}
So far, we have discussed the criteria used by humans for evaluating NLG systems. However, as established earlier, procuring such ratings on a large scale every time a new system is proposed or modified is expensive, tedious and time consuming. Hence, automatic evaluation metrics have become popular. Over the years, many automatic metrics have been proposed, some task-specific and some task-agnostic. Before describing these metrics, we first present a taxonomy of these metrics. To do so, we introduce some notation to refer to the context (or input), reference (or ground-truth) and the hypothesis (or the generated output) which is to be evaluated. 
The context varies from one task to another and could be a document, passage, image, graph, \textit{etc}. Additionally, the expected output text is referred to by a specific term in relation to the context. For example, in the case of translation, the context is the source language sentence which is to be translated. The expected output is referred to as the ``translation'' of the source sentence into the target language. We list the various inputs and outputs for each of the NLG tasks in table \ref{tab:cont_and_ref_forms}. 
\begin{table}[h]
\centering
    \begin{tabular}{l|l|l}
         NLG task & Context & Reference and Hypothesis \\
         \hline
         Machine Translation (MT) & Source language sentence & Translation\\
         Abstractive Summarization (AS) & Document & Summary\\
         Question Answering (QA) & Question + Background info (Passage, Image, etc) & Answer\\
         Question Generation (QG) & Passage, Knowledge base, Image & Question\\
         Dialogue Generation (DG) & Conversation history & Response\\
         Image captioning (IC) & Image & Caption\\
         Data to Text (D2T) & Semi-structured data (Tables) & Description\\
         \hline
    \end{tabular}
    \caption{Context and reference/hypothesis forms for each NLG task}
    \label{tab:cont_and_ref_forms}
\end{table}

In the following sections discussing the existing automatic metrics, we use the generic terms, context, reference, and hypothesis denoted by $c$, $r$, and $p$ respectively. 
The reference and hypothesis would be a sequence of words and we denote the lengths of these sequences as $|r|$ and $|p|$ respectively.
In case there are multiple reference sentences available for one context, we represent the set of references as $R$. Text-based contexts (sentences, documents, passages) also contain a sequence of words and we refer to the length of this sequence as $|c|$. In the case of `conversation history', there could be additional delimiters to mark the end of an utterance, and distinguish between the speakers involved in the conversation. Images are represented as matrices or multidimensional arrays. Tables are expressed as a set of records or tuples of the form, \textit{(entity, attribute, value)}. The notations for any such special/additional elements are introduced as and when required.\\

Given the above definitions, we classify the existing metrics using the taxonomy summarized in Figure \ref{fig:taxo}. We start with 2 broad categories: (i) Context-free metrics and (ii) Context-dependent metrics. Context-free metrics do not consider the context while judging the appropriateness of the hypothesis. In other words, they only check the similarity between the hypothesis and the given set of references. This makes them task-agnostic and easier to adopt for a wider variety of NLG tasks (as irrespective of the task, the reference and hypothesis would just be a sequence of words that need to be compared). Table \ref{tab:auto_metric_adoption} depicts the NLG tasks for which each of the automatic metrics were proposed and/or adopted for.  On the other hand, context-dependent metrics also consider the context while judging the appropriateness of the hypothesis. They are typically proposed for a specific task and adopting them for other tasks would require some tweaks. For example, a context-dependent metric proposed for MT would take the source sentence as input and hence it cannot directly be adopted for the task of image captioning or data-to-text generation where the source would be an image or a table. We thus categorize context-dependent metrics based on the original tasks for which they were proposed. 
We further classify the metrics based on the techniques they use. For example, some metrics are trained using human annotation data whereas some other metrics do not require any training and simply use a fixed set of heuristics. The untrained metrics can be further classified based on whether they operate on words, characters, or word embeddings. Similarly, the trained metrics could use other metrics/heuristics as the input features or be trained in an end-to-end fashion using the representations of the reference, hypothesis, and context. For learning the parameters of a trained metric, various machine learning techniques such as linear regression, SVMs, deep neural networks, \textit{etc.}, can be used. This trained/untrained categorization is applicable to both the context-free and context-dependent metrics. However, we find that currently most of the context-dependent metrics are trained, with only a handful of untrained metrics. 
With this taxonomy 
we discuss the various context-free and context-dependent metrics in the next 2 sections.
\input{taxonomy1}
\begin{table}[h]
    \centering
    \begin{tabular}{c|c|c|c|c|c|c|c}
         Metric & MT & AS & DG & IC  & QA & D2T & QG\\
         \hline
         \multicolumn{8}{c}{Context-free metrics}\\
         \hline
         BLEU & \checkmark & * & * & * & * & * & *\\
         NIST & \checkmark & * & * & * & * & *& *\\
         METEOR & \checkmark & * & * & * & * & *& *\\
         ROUGE & * & \checkmark & * & * & * & *& *\\
         GTM & \checkmark & * &  &  & * & &\\
         CIDEr &  &  &  & \checkmark & & &\\
         SPICE &  &  &  & \checkmark &  & &\\
         SPIDer &  &  &  & \checkmark & & &\\
         WER-family & \checkmark &  &  &  &  & &\\
         chrF & \checkmark & * &  & * &  & &\\
         Vector Extrema & * & * & * & * & * & *& \\
         Vector Averaging & * & * & * & * & * & *& \\
         WMD & * & * &  & * &  & &\\
         BERTr & * &  &  &  &  & &\\
         BERTscore & \checkmark &  & * & \checkmark & * & &\\
         MoverScore & \checkmark & \checkmark &  & \checkmark &  & \checkmark &\\
         BEER & \checkmark &  &  &  &  & & \\
         BLEND & \checkmark &  &  &  &  &  &\\
         Q-metrics &  &  &  &  &  & & \checkmark\\
         Composite metrics &  &  &  & \checkmark & & &\\
         SIMILE & \checkmark &  &  &  &  &  &\\
         ESIM & \checkmark &  &  &  &  &  &\\
         RUSE & \checkmark &  &  &  &  &  &\\
         BERT for MTE & \checkmark &  &  &  &  &  &\\
         BLEURT & \checkmark &  &  &  &  & \checkmark &\\
         NUBIA & \checkmark &  &  & \checkmark &  & & \\
         \hline
         \multicolumn{8}{c}{Context-dependent metrics}\\
         \hline
         ROUGE-C &  & \checkmark &  & &  & & \\
         PARENT &  &  &  & &  & \checkmark & \\
         LEIC &  &  &  & \checkmark &  & &\\
         ADEM &  &  & \checkmark &  &  & &\\
         RUBER &  &  & \checkmark &  &  & &\\
         SSREM &  &  & \checkmark &  &  & &\\
         RUBER with BERT embeddings &  &  & \checkmark &  &  & &\\
         MaUde &  &  & \checkmark &  &  & &\\
         RoBERTa-eval &  &  & \checkmark &  &  & &\\
         \hline
    \end{tabular}
    \caption{Automatic metrics proposed (\checkmark) and adopted (*) for various NLG tasks }
    \label{tab:auto_metric_adoption}
\end{table}

\section{Context-Free metrics}
\label{sec:context_free_metrics}
In this section, we discuss the various context-free metrics \textit{i.e.}, metrics which do not take the input context into consideration during evaluation. Context-free metrics evaluate a hypothesis by comparing it with the set of available references. Context-free metrics can be broadly categorized into two categories (i) Untrained Metrics: metrics that use pre-defined heuristic-based features such as n-gram precision, recall, and hence are not learnable (ii) Trained Metrics: metrics which contain learnable components that are trained specifically for the task of automatic evaluation. We discuss the different metrics under these two categories in the next two subsections. 


\subsection{Untrained metrics}
Untrained metrics can be further classified into three categories based on the type of features they use \textit{viz. } (i) Word-based (ii) Character-based (iii) Embedding-based. We discuss these in detail below.

\subsubsection{\textbf{Word-based metrics}} Word-based metrics typically treat the hypothesis and the reference as a bag of words or $n$-grams ($n$ contiguous words). They then assign a score to the hypothesis based on the word or $n$-gram overlap between the hypothesis and the reference. Alternatively, some other metrics assign a score to the hypothesis based on the number of word edits required to make the hypothesis similar to the reference. Most of the early evaluation metrics such as BLEU, NIST, METEOR, etc. are all word-based metrics. Given their simplicity and ease of use, these metrics have been widely adopted for many NLG tasks. 
\\
\\
\textbf{BLEU} (Bilingual Evaluation Understudy \cite{DBLP:conf/acl/PapineniRWZ02_bleu}): This was the among the first and most popular metrics proposed for automatic evaluation of MT systems. 
It is a precision-based metric that computes the $n$-gram overlap between the reference and the hypothesis. In particular, BLUE is the ratio of the number of overlapping $n$-grams to the total number of $n$-grams in the hypothesis. To be precise, the numerator contains the sum of the overlapping $n$-grams across all the hypotheses (\textit{i.e.}, all the test instances) and the denominator contains the sum of the total $n$-grams across all the hypotheses (\textit{i.e.}, all the test instances). This precision is computed separately for different values of $n$ as shown below. 
\begin{align*}
    precision_n  = \frac{\sum\limits_{p \in \text{hypotheses}} \sum\limits_{\text{$n$-gram} \in p} Count_{clip}(\text{$n$-gram})}{\sum\limits_{p \in \text{hypotheses}} \sum\limits_{\text{$n$-gram} \in p} Count(\text{$n$-gram})}
\end{align*}
where $Count_{clip}(\text{$n$-gram})$ is clipped by the maximum number of times the given $n$-gram appears in any one of the corresponding reference sentences. \newA{For example, if a particular $n$-gram appears thrice in the hypothesis, but twice in one reference and once in another reference in a multireference setting, then we want to consider the matched $n$-gram count as 2 and not as 3.} More precisely,
    \begin{equation*}
    Count_{clip}(\text{$n$-gram}) = \min\Big(\text{matched $n$-gram count, } \max_{r \in R} (\text{$n$-gram count in r} )\Big)
    \end{equation*}
Note that we refer to an $n$-gram in the hypothesis which overlaps with an $n$-gram in the reference as a matched $n$-gram.

\MK{the above formula is not clear to me. Let's discuss. I think we need to distinguish between what happens when there is a single reference v/s what happens when there are multiple references. \Ana{Does the above example help?}}

    
Once the above precision is computed for different values of $n$, a final $BLEU$-$N$ score is computed as a weighted combination of all the $precision_n$ scores, $ n = 1,..,N$. In the original paper, $BLEU$-$N$  was computed as the geometric mean of all the $precision_n$ scores, $ n = 1,..,N$.
Since precision depends only on the length of the hypothesis and not on the length of the sentence, an NLG system can exploit the metric and acquire high scores by producing only a few matching or common words/$n$-grams as the hypothesis. To discourage such short meaningless hypothesis, a brevity penalty term, BP, is added to the formula:
\begin{align*}
    BP = 
    \begin{cases}
     1, & \text{if } |p|>|r| \\
     e^{\big(1-\frac{|r|}{|p|}\big)} & \text{otherwise}
    \end{cases}
\end{align*}
The final formula popularly used today is 
\begin{align*}
    BLEU\text{-}N= BP\cdot exp\bigg(\sum_{n=1}^N W_n\log precision_n \bigg)
\end{align*}
where $W_n$ are the weights of the different $n$-gram precisions, such that $\sum_{n=1}^N W_n = 1$. (Usually each $W_n$ is set to $\frac{1}{N}$.)\\

Since each $precision_n$ is summed over all the hypotheses, BLEU is called a corpus-level metric, \textit{i.e.}, BLEU gives a score over the entire corpus (as opposed to scoring individual sentences and then taking an average). Over the years, several variants of BLEU have been proposed. \textbf{SentBLEU} is a smoothed version of BLEU that has been shown to correlate better with human judgements at the \newA{sentence-level}. \MK{what does segment mean? sentence? \Ana{Yeah, I've changed it. SentBLEU is sometimes expanded as sentenceBLEU and is used to report sentence-level scores. However, in WMT task, they distinguish `system-level' from `segment-level', rather than sentence-level. I'm not entirely sure why. Update: In some of the older papers, I found this: ` segments  should  be no  less  than  one  sentence  in length'.}} Recently, there was a push for standardizing BLEU \cite{DBLP:conf/wmt/Post18_bleu_std_sacrebleu} by fixing the tokenization and normalization scheme to the one used by the annual Conference on Machine Translation (WMT). This standardized version is referred to as \textbf{sacreBLEU}. 
Discriminative BLEU or \textbf{$\Delta$-BLEU} \cite{DBLP:conf/acl/GalleyBSJAQMGD15_deltableu} uses human annotations on a scale [-1,+1] to add weights to multireference BLEU. The aim is to reward the $n$-gram matches between the hypothesis and the good references, and penalize the $n$-grams that only match with the low-rated references. Thus, each $n$-gram is weighted by the highest scoring reference in which it occurs and this weight can sometimes be negative.\\
\\
\textbf{NIST\footnote{The name NIST comes from the organization, ``US National Institute of Standards and Technology".}} \cite{10.5555/1289189.1289273_nist}: This metric can be thought of as a variant of BLEU which weighs each matched $n$-gram based on its information gain. The information gain for an $n$-gram made up of words $w_1,..,w_n$, is computed \newA{over the set of reference  translations,} as
\begin{align*}
    Info(\text{$n$-gram}) = Info(w_1, ..., w_n) = \log_{2} \frac{\text{\# of occurrences of } w_1,...,w_{n-1}}{\text{\# of occurrences of } w_1,...,w_n }
\end{align*}
\MK{in the denominator do we consider only references sentences or all sentences in the corpus? What about numerator? \Ana{There are not many details given about this. As I understand, yes both numerator and denominator are computed only over (all the) reference sentences (esp since it came in the context of translation, I don't think they had other options). Anyway, I have added this information in the text rather than the formula now, almost exactly as presented in the paper. I now removed ``reference sentences" from the denominator as in the original equation. }}
The idea is to give more credit if a matched $n$-gram is rare and less credit if a matched $n$-gram is common. This also reduces the chance of gaming the metric by producing trivial $n$-grams. \newA{The authors further forgo the use of geometric mean to combine the different $precision_n$ scores which makes the contribution of $n$-grams of different length difficult to interpret.}
\MK{the portion marked in square brackets is not very clear to me. Please word it better. Use more sentences to explain it, if required. \Ana{Yeah, this was one part that wasn't clear to me in the paper. I've rephrased it above in a slightly different way. The original point I had written is commented out below. For the exact words by the authors, see section 4, 1st point here: https://dl.acm.org/doi/pdf/10.5555/1289189.1289273}}
In addition to these changes, NIST also modifies the brevity penalty term in order to reduce the impact of small variations in hypothesis length $p$ on the score. To easily compare all these changes in NIST (as a variant of BLEU), note that BLEU formula can be written as follows by expanding the penalty term:
\begin{align*}
    BLEU\text{-}N= \exp\bigg(\sum_{n=1}^N W_n\log precision_n \bigg) \cdot \exp \bigg(\min\Big(1-\frac{|r|}{|p|},0\Big)\bigg)
\end{align*}
\begin{align*}
    NIST = \sum_{n=1}^N \bigg\{\frac{\sum_{\text{all $n$-grams that match}}Info(\text{$n$-gram})}{\sum_{n\text{-}gram \in hypotheses}(1)} \bigg\} \cdot \exp \bigg(\beta \log^2 \Big[\min\Big(\frac{|p|}{|\bar{r}|},1\Big) \Big] \bigg)
\end{align*}
where $\beta $ is chosen to make brevity penalty factor = $0.5$ when the number of words in the hypothesis is $2/3^{rds}$ of the average number of words in the reference, and $|\bar{r}|$ is the average number of words in a reference (averaged over all the references).\\
\MK{Is the above formula taken as it is from the original paper? If not, is there a better way of writing it? \Ana{The formula was taken from the original paper.} In the numerator, by cooccur do you mean match? \Ana{Yes, I've replaced it now.} Shouldn't we be dividing by $N$ to take average? \Ana{This is a good catch. Sorry, I did not write that detail correctly in text last time. While talking about discarding the geometric mean, the authors say ``an alternative would be to use an arithmetic average of N-gram counts rather than a geometric average" (over the precisions) -> is missing. } Unlike the BLEU-N formula, here it is not clear that we are using precision\_n for the computation. Can we write this formula in a way that it is easy to compare with the BLEU formula?\Ana{Does the above presentation work? Otherwise I think we formulate it as info-weighted $precision_n$ terms in the numerator divided by total number of $n$-grams.}}
\\
\textbf{GTM} (General Text Matcher) :  \citet{article_general_text_matcher_gtm} observe that systems can game a metric by increasing the precision or recall individually even through bad generations. The authors hence suggest that a good metric should use a combination of precision and recall such, as F-measure (which is the harmonic mean of precision and recall).
\newA{Towards this end they propose `GTM', an F-Score based metric, with greater weights for contiguous word sequences matched between the hypothesis and reference. A `\textit{matching}' is defined as a mapping of words between the hypothesis and the reference, based on their surface-forms, such that no two words of the hypothesis are mapped to the same word in the reference and vice versa. In order to assign higher weights to contiguous matching sequences termed ``runs", weights are computed for each run as the square of the run length. Note that length of a run could also be $1$ for an isolated word match, more generally it is bound to be between $0$ and $\min(|p|,|r|)$. The hypothesis and reference could have multiple possible matchings with different number of runs of various lengths. The size of a matching, \textit{i.e., match size} of $M$ is computed using the weights of its constituent runs as follows:
\begin{align*}
    size(M) = \mysqrt{-1}{4}{q}{\sum_{run \in M}length(run)^q}
\end{align*}
where higher values of $q$ more heavily weight longer runs. By comparing the match sizes, a matching with the maximum match size (MMS) is selected. In practice, since finding the MMS is NP-hard for $q>1$, GTM uses a greedy approximation where the largest non-conflicting mapped sequences are added iteratively to form the matching (and use its size as MMS). 
Using the approximated MMS, the precision and recall are computed as:}
\begin{gather*}
    \text{(Precision) }P  = \frac{MMS(p,r)}{|p|}\text{ , (Recall) } R = \frac{MMS(p,r)}{|r|} 
    \\
    \text{GTM = F-score} = \frac{2PR}{P+R}
\end{gather*}
\MK{where are we using size(M) in the above formula? What is M (it is not defined anywhere)? It is also not clear how we are using the weights in MMS(p,r) \Ana{I've rewritten most of the content for this metric so that the idea becomes clear now.}}
GTM was proposed for evaluating MT systems and showed higher correlations with human judgements compared to BLEU and NIST (with $q =1$). \\
\\
\textbf{METEOR} (Metric for Evaluation of Translation with Explicit ORdering) : \citet{meteor} point out that there are two major drawbacks of BLEU: (i) it does not take recall into account and (ii) it only allows exact $n$-gram matching. To overcome these drawbacks, they proposed METEOR which is based on F-measure and uses a relaxed matching criteria. In particular, even if a unigram in the hypothesis does not have an exact surface level match with a unigram in the reference but is still equivalent to it (say, is a synonym) then METEOR considers this as a matched unigram. More specifically, it first performs exact word (unigram) mapping, followed by stemmed-word matching, and finally synonym and paraphrase matching. It then computes the F-score using this relaxed matching strategy.\\
\MK{throughout the paper please write all equations between begin\{align*\} as opposed to using \$\$}
\begin{gather*}
    P(Precision)=\frac{\# mapped\_unigrams}{\#unigrams\_in\_candidate} \text{ , } R(Recall)=\frac{\# mapped\_unigrams}{\#unigrams\_in\_reference} 
    \\
    F score = \frac{10PR}{R + 9P}
\end{gather*}
Since METEOR only considers unigram matches (as opposed to $n$-gram matches), it seeks to reward longer contiguous matches using a penalty term \newA{ known as `fragmentation penalty'. To compute this, `chunks' of matches are identified in the hypothesis, where contiguous hypothesis unigrams that are mapped to contiguous unigrams in a reference can be grouped together into one chunk. Therefore longer $n$-gram matches lead to fewer number of chunks, and the limiting case of one chunk occurs if there is a complete match between the hypothesis and reference. On the other hand, if there are no bigram or longer matches, the number of chunks will be the same as the number of unigrams. The fewest possible number of chunks a hypothesis can have is used to compute the fragmentation penalty used in METEOR as:} 
\begin{align*}
    \text{Penalty} = 0.5*\Bigg[\frac{\#chunks}{\#unigrams\_matched}\Bigg]^3 \\
    \text{METEOR Score} = F score*(1-Penalty)
\end{align*}
\MK{what is \# chunks in the above equation \Ana{Added above}}    
Similar to BLEU, METEOR also has a few variants. \newA{ For example, \citet{DBLP:conf/wmt/DenkowskiL10_meteor_next} propose \textbf{METEOR-NEXT} to compute weighted precision and recall by assigning weights to the different matching conditions or the \textit{`matchers'} used (\textit{viz.}, exact, stem, synonym and paraphrase matching):
\begin{align*}
     P=\frac{\sum_{i \in |\{matchers\}|}w_i.m_i(p)}{|p|} \text{ , } R=\frac{\sum_{i \in |\{matchers\}| }w_i.m_i(r)}{|r|} 
\end{align*}
where $m_i(p)$ and $m_i(r)$ represent the counts of the mapped words identified by that particular matcher $m_i$ in the hypothesis and reference respectively, and $w_i$ is the corresponding weight. The parameterized F-score is calculated as
\begin{align*}
    F score = \frac{PR}{\alpha.P+(1-\alpha).R}
\end{align*}
Further building on this variant, \citet{DBLP:conf/wmt/DenkowskiL14_meteor_universal} observe that METEOR uses language specific resources (for stemming and matching synonyms) and propose \textbf{METEOR Universal} that generalizes across languages by automatically building function-word lists and paraphrase lists using parallel text in different languages. With these lists, they define weighted precision and recall similar to METEOR-NEXT that additionally has the flexibility to weigh the content words and function words differently:
\begin{align*}
    P=\frac{\sum_{i}w_i.(\delta . m_i(p_c)+ (1-\delta).m_i(p_f) }{\delta.|p_c| + (1-\delta).|p_f|} \text{ , } R=\frac{\sum_{i}w_i.(\delta . m_i(r_c)+ (1-\delta).m_i(r_f)}{\delta.|r_c| + (1-\delta).|r_f|} 
\end{align*}
where $p_c$ and $r_c$ denote the content words in hypothesis and reference, while $p_f$ and $r_f$ represent the function words and $\delta, w_i^s$ are parameters. In order to have a language-agnostic formula, all the parameters are tuned to encode general human preferences that were empirically observed to be common across languages, such as, preferring recall over precision, word choice over word order, correct translation of content words over function words, etc.}
\MK{what are these parameters? the above equations do not contain any parameters which can be tuned. \Ana{Added above}}   \textbf{METEOR++ }\cite{guo-etal-2018-meteorpp} additionally incorporates ``copy-words" specially into the metric, to deal with the words that have a high-probability of remaining the same throughout all paraphrases of a sentence. These could be named-entities or words like \textit{traffic, government, earthquake} which do not have many synonyms. Based on these, METEOR++ aims to capture whether the hypothesis is incomplete (with missing copy words) or inconsistent (with spurious copy-words). \textbf{METEOR++2.0} \cite{DBLP:conf/wmt/GuoH19_meteorpp2} also considers syntactic level paraphrases which are not necessarily contiguous (such as ``not only ... but also ... '') rather than considering only lexical-level paraphrases of consecutive $n$-grams.\\
\\
\textbf{ROUGE} (Recall-Oriented Understudy for Gisting Evaluation \cite{rouge}) : 
ROUGE metric includes a set of variants: ROUGE-N, ROUGE-L, ROUGE-W, and ROUGE-S. ROUGE-N is similar to BLEU-N in counting the $n$-gram matches between the hypothesis and reference, however, it a recall-based measure unlike BLEU which is precision-based.
\begin{align*}
    \text{ROUGE-N}  = \frac{\sum\limits_{s_r \in \text{references}} \sum\limits_{\text{$n$-gram} \in s_r} Count_{match}(\text{$n$-gram})}{\sum\limits_{s_r \in \text{references}} \sum\limits_{\text{$n$-gram} \in s_r} Count(\text{$n$-gram})}
\end{align*}
ROUGE-L measures the longest common subsequence (LCS) between a pair of sentences. Note that a sequence $Z = [z_1, z_2, ...., z_n]$ is called a subsequence of another sequence $X = [x_1, x_2, ..., x_m]$ if there exists a strictly increasing sequence $[i_1, i_2, ... i_n]$ of indices of X such that $x_{i_j} = z_j$ for all $j=1,2,...,n$ \cite{cormen2009introduction}. The \textit{longest common subsequence}, $LCS(p,r)$ is the common subsequence in $p$ and $r$ with maximum length. ROUGE-L is a F-measure where the precision and recall are computed using the the length of the LCS:
\begin{gather*}
    P_{lcs} = \frac{|LCS(p,r)|}{\# words\_in\_hypothesis} \text{ , } R_{lcs} = \frac{|LCS(p,r)|}{\# words\_in\_reference}
    \\
    \text{ROUGE-L} = F_{lcs} = \frac{(1+\beta^2)R_{lcs}P_{lcs}}{R_{lcs} + \beta^2 P_{lcs}}
\end{gather*}
Note that ROUGE-L does not check for consecutiveness of the matches as long as the word order is the same. It hence cannot differentiate between hypotheses that could have different semantic implications, as long as they have the same LCS even with different spatial positions of the words w.r.t the reference. ROUGE-W addresses this by using a weighted LCS matching that adds a \textit{gap penalty} to reduce weight on each non-consecutive match.\\
ROUGE-S uses skip-bigram co-occurrence statistics to measure the similarity of the hypothesis and reference. Skip-bigrams are pairs of words in the same sentence order, with arbitrary words in between. ROUGE-S is also computed as an F-score similar to ROUGE-L.\\
ROUGE variants were originally proposed for evaluating automatic summarization, but have been adopted for evaluation of other NLG tasks.\\  
\\
\textbf{CIDEr} (Consensus-based Image Description Evaluation \cite{DBLP:conf/cvpr/VedantamZP15_CIDEr}) : CIDEr weighs each $n$-gram in a sentence based on its frequency in the corpus and in the reference set of the particular instance, using TF-IDF (term-frequency and inverse-document-frequency). It was first proposed in the context of image captioning where each image is accompanied by multiple reference captions. It is based on the premise that $n$-grams that are relevant to an image would occur frequently in its set of reference captions. 
However, $n$-grams that appear frequently in the entire dataset (\textit{i.e.}, in the reference captions of different images) are less likely to be informative/relevant and hence they are assigned a lower weight using inverse-document-frequency (IDF) term. To be more precise, the TF-IDF weight, $g_{n_k}(s)$, for each $n$-gram $k$ in caption $s_i$ are computed as follows: 
\begin{align*}
    g_{n_k}(s) = \frac{t_k(s)}{\sum_{l \in V_n}t_l(s)} \log \Bigg( \frac{|I|}{\sum_{i \in I}\min(1,\sum_{r \in R_i}t_k(r))} \Bigg)
\end{align*}
where $V_n$ is the vocabulary of all $n$-grams, $g_{n_k}$ refers to the weight assigned to an $n$-gram denoted by $k$, $t_k(s)$ is the number of times $k$ appears in $s$, $I$ is the set of all images, $R_i$ corresponds to the set of references for image $i$.\\
\Ana{CIDEr paper seems to have conflict between text and formula for the TF term. Shouldn't the denominator of the first term sum over all the references for it to be consistent with what the authors have written in text?}
CIDEr first stems the words in hypothesis and references and represents each sentence as a set of $n$-grams. It then calculates weights for each $n$-gram using TF-IDF as explained above.
\MK{How did we suddenly transition to cosine similarity? \Ana{Changed content above. The TF-IDF weights are combined.}} 
Using these TF-IDF weights of all the $n$-grams of length $n$, vectors $g_n(s)$ are formed for each caption $s$. $CIDEr_n$ is calculated as the average cosine similarity between hypothesis and references:
\begin{align*}
    CIDEr_n(p, R) = \frac{1}{|R|}\sum_{r \in R}\frac{g_n(p).g_n(r)}{||g_n(p)||~||g_n(r)||}
\end{align*}
Final CIDEr score is the weighted average of $CIDEr_n$ for $n = 1,2,3,4$:
$$ CIDEr(p,R) = \sum_{n=1}^N W_n CIDEr_n(p,R) $$
where the weights are uniform $W_n = \frac{1}{N}$ and N is set to 4.\\
\\
\textbf{SPICE} (Semantic Propositional Image Caption Evaluation \cite{DBLP:conf/eccv/AndersonFJG16_SPICE} ): In the context of image captioning, \citet{DBLP:conf/eccv/AndersonFJG16_SPICE} suggest that instead of focusing on $n$-gram similarity, more importance should be given to the semantic propositions implied by the text. To this end, they propose SPICE which uses `scene-graphs' to represent semantic propositional content. In particular, they parse the sentences into semantic tokens such as object classes $C$, relation types $R$ and attribute types $A$. Formally, a sentence $s$ is parsed into a scene-graph $G(s)$ as: 
$$ G(s) = <O(s), E(s), K(s)>$$
where $O(s) \subseteq C$ is the set of object mentions in $s$, $E(s) \subseteq O(s) \times R \times O(s)$ is the set of hyperedges representing relations between objects, and $K(s)\subseteq O(s) \times A$ is the set of attributes associated with objects. The hypothesis and references are converted into scene graphs and the SPICE score is computed as the F1-score between the scene-graph tuples of the proposed sentence and all reference sentences. For matching the tuples, SPICE also considers synonyms from WordNet\cite{DBLP:conf/aaai/PedersenPM04_wordnet} similar to METEOR \cite{meteor}. One issue with SPICE is that it depends heavily on the quality of parsing. Further, the authors note that SPICE neglects fluency assuming that the sentences are well-formed. It is thus possible that SPICE would assign a high score to captions that contain only objects, attributes and relations, but are grammatically incorrect.\\
\\
\textbf{SPIDEr} \cite{DBLP:conf/iccv/LiuZYG017_spider} \footnote{The name is a fusion of `SPICE' and `CIDEr'} : This metric is a linear weighted combination of SPICE and CIDEr. The motivation is to combine the benefits of semantic faithfulness of the SPICE score and syntactic fluency captured by the CIDEr score. Based on initial experiments, the authors use equal weights for SPICE and CIDEr.\\
\\
\textbf{WER} (Word Error Rate): There is a family of WER-based metrics which measure the edit distance $d(c,r)$, \textit{i.e.}, the number of insertions, deletions, substitutions  and, possibly, transpositions required to transform the candidate into the reference string. Word Edit Rate (WER) was first adopted for text evaluation from speech evaluation by \citet{su-etal-1992-new_wer_to_mt_from_speech} in 1992. Since then, several variants and enhancements have been proposed, as discussed below. The original formula is is based on the fraction of word edits as given below: 
$$ WER = \frac{\# of substitutions+insertions+deletions}{reference\ length} $$
Since WER relies heavily on the reference sentence,  \citet{DBLP:conf/lrec/NiessenOLN00_multiWER} propose enhanced WER that takes into account multiple references. Another issue with WER is that it penalizes different word order heavily since each ``misplaced" word triggers a deletion operation followed by an insertion operation, when infact the hypothesis could still be valid even with a different word order.
To account for this, \textbf{TER} (Translation Edit Rate \cite{Snover06astudy_ter}) adds a shifting action/block movement as an editing step. 
\textbf{ITER} \cite{DBLP:conf/wmt/PanjaN18_iter} is a further improved version of TER. In addition to the basic edit operations in TER (insertion, deletion, substitution and shift), ITER also allows stem matching and uses optimizable edit costs and better normalization. \textbf{PER} \cite{DBLP:conf/interspeech/TillmannVNZS97_per} computes `Position-independent Edit Rate' by identifying the alignments/matching words in both sentences. Then depending on whether the proposed sentence is shorter or longer than the reference, the remaining words are counted as insertions or deletions.
\textbf{CDER} \cite{DBLP:conf/eacl/LeuschUN06_cder} models block reordering as a unit edit operation to off-set unnecessary costs in shifting words individually.

\subsubsection{\textbf{Character-based metrics}}

The metrics that we have discussed so far, operate at the word level. In this subsection, we discuss evaluation metrics which operate at the character level. These metrics usually do not require tokenization to identify the tokens in the sentence, and directly work on the reference and hypothesis strings. Note that some of these metrics additionally enlist the help of word-level information. The main motivation for using character-based metrics is their improved performance in evaluating morphologically rich languages \cite{DBLP:conf/wmt/WangPRN16_characTER,DBLP:conf/wmt/Popovic15_chrF}. \\
\\ 
\textbf{characTER} by \citet{DBLP:conf/wmt/WangPRN16_characTER} is a character-level metric inspired by the Translation Edit Rate (TER) metric discussed above.  CharacTER first performs shift edits at the word level, using a relaxed matching criteria where a word in the hypothesis is considered to match a word in the reference if the character based edit distance between them is below a threshold value. Then the shifted hypothesis sequence and the reference are split into characters and the Levenshtein distance between them is calculated. \newA{Additionally, since normalizing by reference length (as done in TER) does not take the hypothesis length into account, characTER uses the length of hypothesis for normalizing the edit distance. This normalization is empirically shown to correlate better with human judgements.}\\
\\
\textbf{EED} (Extended Edit Distance) \citet{stanchev-etal-2019-eed}: This metric is inspired by CDER and extends the conventional edit operations (insertions, deletions and substitutions) to include a jump operation, but at the character level. \newA{Jumps provide an opportunity to continue the edit distance computation from a different point. This would be useful if, for example, the hypothesis has a different word order than the reference. However, in order to avoid jumps in the middle of a word, this operation is permitted only on blank space characters (\textit{i.e.}, disallowing inter-word jumps).} \MK{what does this mean? is this character level or word level jump? \Ana{Rephrased above content}} Further, if any of the hypothesis characters are aligned to multiple characters in the reference or not aligned at all, their counts are added to form a coverage-penalty term $v$. EED is defined as:
\begin{align*}
    EED = \min \Bigg(\frac{(e+\alpha.j) + \rho.v}{|r| + \rho.v} , 1 \Bigg)
\end{align*}
where $e$ denotes the cost of the conventional edit operations with a uniform cost of $1$ for insertion and substitution and $0.2$ for deletion. $j$ is the number of jump operations, $\alpha$ and $\rho$ are parameters optimised to correlate well with human judgements on WMT17 and WMT18 \cite{DBLP:conf/wmt/BojarGK17_wmt-17-results,wmt-2018-results}. 
Note that the coverage-penalty term is also added to the length of the reference in the denominator, \textit{i.e.}, the normalisation term, to naturally keep the score between [0,1] and reduce the number of times the min function chooses the value $1$ over the result of the formula.\\
\MK{Suddenly we have introduced a coverage penalty term? Use more sentences to explain this. Alternatively providing a formula may help. \Ana{Added formula and rephrased content. Should we can comment out the last sentence about the why the penalty term is in the denominator?}}.
\MK{without a formula it is hard to understand to what are you referring to when you say denominator.} \\
\\
\textbf{chrF} \cite{DBLP:conf/wmt/Popovic15_chrF}: This metric compares character $n$-grams in the reference and candidate sentences, instead of matching word $n$-grams as done in BLEU, ROUGE, etc. \newA{The precision and recall are computed over the character $n$-grams for various values of $n$ (upto 6) and are combined using arithmetic averaging to get the overall precision ($chrP$) and recall ($chrR$) respectively. In other words, $chrP$ represents the percentage of matched character $n$-grams present in the hypothesis and $chrR$ represents the percentage of character $n$-grams in the reference which are also present in the hypothesis, where $n \in [1,2,..,6]$.} \MK{do yo mean to say that these averages are computed over all values of $n$? If not, it is not clear what the arithmetic average is? \Ana{Yes, I have clarified that now} The description is a bit rushed. What is the value of $n$ here? Use a couple of more sentences to explain clearly. \Ana{Modified the above description slightly}} 
The final chrF score is then computed as:
\begin{align*}
    chrF_\beta = (1+\beta^2) \frac{chrP.chrR}{\beta^2.chrP + chrR}
\end{align*}
where the value of $\beta$ indicates that recall is given $\beta$ times more weightage than precision. 
chrF was initially proposed for evaluating MT systems but has been adopted for other tasks such as image captioning and summarization as well. \citet{DBLP:conf/wmt/Popovic17_chrFpp} propose enhanced versions of chrF, which also contain word $n$-grams in addition to character $n$-grams. These include \textbf{chrF+} which also considers word unigrams and \textbf{chrF++} which considers word unigrams and bigrams in addition to character $n$-grams.

\subsubsection{\textbf{Embedding based metrics}}
The word/character based metrics discussed above, rely largely on surface level matches (although a couple of them do consider synonyms). As a result, they often ignore semantic similarities between words. For example, the words `canine' and `dog' are related, and are synonyms in some contexts. Similarly, the words `cat' and `dog', although not synonymous, are closer (by virtue of being pet animals) than say, `dog' and `boat'. Such similarities are better captured by word embeddings such as as Word2Vec \cite{DBLP:journals/corr/abs-1301-3781_word2vec}, GloVe \cite{DBLP:conf/emnlp/PenningtonSM14_glove}, \textit{etc.}, which are trained on large corpora and capture distributional similarity between words. Thus, an alternative to matching words is to compare the similarity between the embeddings of words in the hypothesis and the reference(s). We discuss such word embedding based metrics in this subsection. 
In all the discussion that follows, we represent the embedding of a word $w$ as $\overrightarrow{w}$.\\
\\
\textbf{Greedy Matching} \cite{greedymatch}: This metric considers each token in the reference and greedily matches it to the closest token in the hypothesis based on the cosine similarity between the embeddings of the tokens. The aggregate score is obtained by averaging across all the tokens in the reference. However, this greedy approach makes this score direction-dependent, and hence the process is repeated in the reverse direction (\textit{i.e.}, greedily match each hypothesis token with the reference tokens) to ensure that the metric is symmetric. The final score given by greedy matching metric (GM) is the average of matching in both directions. 
$$ G(p,r) = \frac{\sum_{w\in r}\max_{\hat{w} \in p} cosine(\overrightarrow{w},\overrightarrow{\hat{w}})}{|r|} $$
$$ GM = \frac{G(p,r)+G(r,p)}{2} $$
\\
\textbf{Embedding Average metric} \cite{landauer1997solution_vector_averaging1} : Instead of computing a score for the hypothesis by comparing the embeddings of the words/tokens in the hypothesis and the reference, one could directly compute and compare the embeddings of the sentences involved (\textit{i.e.}, the hypothesis sentence and the reference sentence). The Vector Averaging or Embedding Average metric does exactly this by first computing a sentence-level embedding by averaging the word embeddings of all the tokens in the sentence.
$$ \overrightarrow{s} = \frac{\sum_{w\in s}\overrightarrow{w}}{|s|} $$

The score for a given hypothesis, $EA$, is then computed as the cosine similarity between the embedding of the reference ($\overrightarrow{r}$) and the embedding of the hypothesis ($\overrightarrow{p}$).
$$ EA = cosine(\overrightarrow{p},\overrightarrow{r})$$
\\
\textbf{Vector Extrema:} The sentence-level embeddings can alternatively be calculated by using Vector Extrema \cite{vectorextrema}. In this case, a $k$-dimensional sentence embedding is constructed using the $k$-dimensional word embeddings of all the words in the sentence. However, instead of taking an average of the word embeddings, a dimension-wise max/min operation is performed over the word embeddings. In other words, the most extreme value (\textit{i.e.}, the value farthest from $0$) along each dimension is chosen by considering the embeddings corresponding to all the words in the sentence. 
\begin{align*}
    \overrightarrow{s_d} = 
    \begin{cases}
     \max_{w \in s}\overrightarrow{w}_d, & \text{if } \overrightarrow{w}_d>|\min_{w'\in s}\overrightarrow{w'}_d| \\
     \min_{w\in s}\overrightarrow{w}_d & \text{otherwise}
    \end{cases}
\end{align*}
\MK{the above formula doesn't make sense to me. the if and else parts are the same \Ana{extrema value can be negative. I verified this with the formula in paper }}
where $d$ indexes the dimensions of a vector. 
The authors claim that by taking the extreme value along each dimension, we can ignore the common words (which will be pulled towards the origin) and prioritize informative words which will lie further away \newA{from the origin} in the vector space. \MK{is this phrasing correct? is it right to say ``further away in the vector space''? \Ana{should we say ``further away from the origin in the vector space''? If it's between further and farther, both are appropriate in this case.}} The final score assigned to a hypothesis is the cosine similarity between the sentence-level embeddings of the reference and the hypothesis.\\
\\
\textbf{WMD} (Word Mover-Distance) \cite{DBLP:conf/icml/KusnerSKW15_wmd}: This metric was proposed to measure dissimilarity between text documents by computing the minimum cumulative distance between the embeddings of their constituent words. It performs optimal matching rather than greedy matching, based on the Euclidean distance between the word embeddings of the hypothesis and reference words. \newA{Note that an optimal matching might have each word embedding in the hypothesis to be partially mapped to multiple word embeddings in the reference. To model this effectively, the hypothesis and reference are first represented as $n$-dimensional normalized bag-of-words vectors, $\overrightarrow{p}$ and $\overrightarrow{r}$ respectively. 
The number of dimensions, $n$, of the normalized bag-of-words vector of a sentence is given by the vocabulary size, and the value of each dimension represents the normalized occurrence count of the corresponding word from the vocabulary in the sentence. That is, if the $i^{th}$ vocabulary word appears $t_i$ times in a sentence $s$, then $\overrightarrow{s}_i= \frac{t_i}{\sum^n_{j=1}t_j}$. WMD allows any word in $\overrightarrow{p}$ to be transformed into any word in $\overrightarrow{r}$ either in total or in parts, to arrive at the minimum cumulative distance between $\overrightarrow{p}$ and $\overrightarrow{r}$ using the embeddings of the constituent words. Specifically, WMD poses a constraint-optimization problem as follows:
\begin{gather*}
     WMD(p,r) = \min_{T}\sum_{i,j=1}^{n}T_{ij}.\Delta(i, j) 
     \\
     \text{such that } \sum_{j=1}^{n}T_{ij} = \overrightarrow{p}_i \forall i \in \{1,..,n\}, \text{ and } \sum_{i=1}^{n}T_{ij} = \overrightarrow{r}_j \forall j \in \{1,..,n\}
\end{gather*}
where $\Delta(i,j) = ||\overrightarrow{w}_i-\overrightarrow{w}_j||_2$ is the Euclidean distance between the embeddings of the words indexed by $i$ and $j$ in the vocabulary \footnote{For simplicity, we here onward refer to a word indexed at $i$ in the vocabulary as simply word $i$}, $n$ is the vocabulary size and $T$ is a matrix with $T_{ij}$ representing how much of word $i$ in $\overrightarrow{p}$ travels to word $j$ in $\overrightarrow{r}$. The two constraints are to ensure complete transformation of $\overrightarrow{p}$ into $\overrightarrow{r}$. That is, the outgoing (partial) amounts of every word $i$ should sum up to the value in the corresponding dimension in $\overrightarrow{p}$ (\textit{i.e.}, its total amount/count in the hypothesis). Similarly, the incoming amounts of every word $j$ in the reference should sum up to its corresponding value in $\overrightarrow{r}$.}
\MK{the rest of this para needs to be re-written. What does bag-of-words vectors mean? You are using $n$ for vector size as well as vocabulary size? Are $\overrightarrow{p}$ and $\overrightarrow{r}$ sentence embeddings or sets of word embeddings? If its the latter the notation is very confusing given that so far you are referring to $\overrightarrow{p}$ as sentence embedding. Slow down. USe more sentences to explain what you wnat to say. The intuition behind the formula is not at all clear.\Ana{I hope the idea is clear now. Should see if we need to word the sentences in a better manner in the next iteration.}} 

Although initially proposed for document classification, WMD has been favourably adopted for evaluating the task of image captioning \cite{DBLP:conf/eacl/KilickayaEIE17_wmd_ic}. 
WMD has also been adopted for summarization and MT evaluation. However, since WMD is insensitive to word order,  \citet{chow-etal-2019-WMDO} propose a modified version termed \textbf{WMD}$_O$ which additionally introduces a penalty term similar to METEOR's fragmentation penalty. \MK{I don't think we mentioned the term ``fragmentation penalty'' while discussing METEOR \Ana{Now I have clarified the fragmentation penalty in METEOR.}} 
\begin{align*}
    WMD_O = WMD - \delta (\frac{1}{2} - penalty)
\end{align*}
where $\delta$ is a weight parameter that controls how much to penalize a different word ordering. In parallel, \textbf{WE\_WPI} (Word Embedding-based automatic MT evaluation using Word Position Information) \cite{echizenya-etal-2019-word_we_wpi} was proposed which also addresses the word-order issue by using an `align-score' instead of Euclidean distance to match words: 
\begin{align*}
    \Delta(i,j) = align\_score = \overrightarrow{w}_i.\overrightarrow{w}_j \times \Bigg(1.0 - \Big|\frac{pos(i,h)}{|h|} - \frac{pos(j,r)}{|r|}\Big|\ \Bigg)
\end{align*}
where $pos(i,h)$ and $pos(j,r)$ indicate the positions of word $i$ in the hypothesis and word $j$ in the reference respectively, and $\Big|\frac{pos(h_i)}{|h|} - \frac{pos(r_j)}{|r|}\Big|$ gives the relative difference between the word positions. WMD$_O$ and WE\_WPI are currently used only for evaluating MT tasks.\\ 
\\
\textbf{MEANT: } \citet{DBLP:conf/wmt/LoTW12_meant_full} make use of semantic role labelling in order to focus on both the structure and semantics of the sentences. Semantic role labelling, also called shallow semantic parsing, is the process of assigning labels to words or phrases to indicate their role in the sentence, such as doer, receiver or goal of an action, \textit{etc}. This annotation would help answer questions like who did what to whom, leading to better semantic analysis of sentences. In this direction, MEANT was proposed as a weighted combination of F-scores computed over the semantic frames as well as their role fillers to evaluate the ``adequacy" of the hypothesis in representing the meaning of the reference. MEANT first uses a shallow semantic parser on the reference and candidate and aligns the semantic frames using maximum weighted bipartite matching based on lexical similarities (of the predicates). This lexical similarity is computed using word vectors \cite{dagan2000contextual_word_similarity}. It then matches the role fillers in a similar manner, and finally computes the weighted F-score over the matching role labels and role fillers.\\
MEANT was originally proposed as a semi-automatic metric \cite{DBLP:conf/acl/LoW11_meant_semi} before the above fully automatic form. There have also been several variants \cite{DBLP:conf/ssst/LoW12_umeant,DBLP:conf/acl/LoBSW14_xmeant} of MEANT metric that followed over the years, with the latest one being \textbf{MEANT2.0}\cite{DBLP:conf/wmt/Lo17_meant2}. MEANT2.0 weighs the importance of each word by IDF (inverse document frequency) to ensure phrases with more matches for content words than for function words are scored higher. It also modifies the phrasal similarity calculation to aggregate on $n$-gram lexical similarities rather than on the bag-of-words in the phrase, so that the word order is taken into account. \\
\hfill\\
\textbf{Contextualized Embedding based metrics}:
The embedding based metrics discussed above use static word embeddings, \textit{i.e.}, the embeddings of the words are not dependent on the context in which they are used. However, over the past few years, contextualized word embeddings have become popular. Here, the embedding of a word depends on the context in which it is used. 
Some popular examples of such contextualized embeddings include ElMo \cite{DBLP:conf/naacl/PetersNIGCLZ18_elmo} , BERT \cite{DBLP:conf/naacl/DevlinCLT19bert} and XLNet \cite{DBLP:conf/nips/YangDYCSL19_xlnet}. In this subsection, we discuss evaluation metrics which use such contextualized word embeddings.\\
\\
\textbf{YiSi:}  YiSi \cite{DBLP:conf/wmt/Lo19_yisi} is a unified semantic evaluation framework that unifies a suite of metrics, each of which caters to languages with different levels of available resources. YiSi-1 is a metric similar to MEANT2.0, that 
uses contextual word embeddings from BERT rather than word2vec embeddings. 
Additionally, it makes the time-consuming and resource-dependent step of semantic parsing used in MEANT2.0 optional. In particular, \MK{The remainder of this sentence is not clear? How is F-score computed using n-gram similarity? What doe we mean by ``weighted'' here? How do we optionally account for semantic structure? Too much crammed in one sentence. Use more sentences to explain this clearly. } YiSi-1 is an F-score that computes $n$-gram similarity as an aggregate of weighted word embeddings cosine similarity, optionally taking the shallow semantic structure into account. \MK{The next sentence is a very long sentence but not at all informative. Accuracy and cosine similarity are to very different things. How do we jump from one to the other? } YiSi-0 is a degenerate resource-free version which uses the longest common character substring accuracy, instead of word embeddings cosine similarity, to measure the word similarity of the candidate and reference sentences. YiSi-2 is the bilingual version which uses the input sentence and is hence discussed in the next section on context-dependent metrics.\\ 
\\
\textbf{BERTr}: \citet{DBLP:conf/acl/MathurBC19_BERTr_contextualized_embeddings_translation} adopt BERT to obtain the word embeddings and show that using such contextual embeddings with a simple average recall based metric gives competitive results.
The BERTr score is the average recall score over all tokens, using a relaxed version of token matching based on BERT embeddings, \textit{i.e.}, by computing the maximum cosine similarity between the embedding of a reference token $j$ and any token in the hypothesis.
\begin{gather*}
    recall_j = \max_{i\in p} cosine(\overrightarrow{i},\overrightarrow{j})
    \\
     \text{BERTr } = \sum_{j\in r}\frac{recall_j}{|r|}
\end{gather*}
\\
\textbf{BERTscore}: \citet{bertscore} compute cosine similarity of each hypothesis token $j$ with each token $i$ in the reference sentence using contextualized embeddings. They use a greedy matching approach instead of a time-consuming best-case matching approach, and then compute the F1 measure as follows:
\begin{gather*}
    R_{BERT} = \frac{1}{|r|}\sum_{i \in r} \max_{j \in p} \overrightarrow{i}^T\ \overrightarrow{j} \text{ , } P_{BERT} = \frac{1}{|p|}\sum_{j \in p} \max_{i \in r} \overrightarrow{i}^T\ \overrightarrow{j}
    \\
    \text{BERTscore} = F_{BERT} = 2 \frac{P_{BERT}.R_{BERT}}{P_{BERT}+ R_{BERT}}
\end{gather*}
The authors show that this metric correlates better with human judgements for the tasks of image captioning and machine translation.\\
\\
\textbf{MoverScore:} \citet{DBLP:conf/emnlp/ZhaoPLGME19_moverscore} take inspiration from WMD metric to formulate another optimal matching metric named MoverScore, which uses contextualized embeddings to compute the Euclidean distances between words or $n$-grams. In contrast to BERTscore which allows one-to-one hard matching of words, MoverScore allows many-to-one matching as it uses soft/partial alignments, similar to how WMD allows partial matching with word2vec embeddings. \MK{I think we need a formula here. \Ana{The formula is similar to WMD, except that MoverScore uses contextualized embeddings. I've now stated this explicitly. We can add the formula for even more clarity, but I'll keep that pending for a later iteration since the formula will get too elaborate like WMD.}} 
It has been shown to have competitive correlations with human judgements in 4 NLG tasks: machine translation, image captioning, abstractive summarization and data-to-text generation.  \MK{What do you mean by generalise? Do you mean it has better correlations for all these tasks? If so, let's just say that.\Ana{Changed}} 

\subsection{Trained metrics}
Evaluation metrics which contain learnable components that are specifically trained for the task of automatic evaluation of NLG systems are categorized as trained metrics. Trained metrics can be further categorized into two classes: (i) Feature-based: metrics which are trained using pre-computed heuristic based features such as $n$-gram precision, recall as input. (ii) End-to-End: metrics which are directly trained using the hypothesis and reference sentences. We shall discuss these two categories in detail in the next two subsections. 


\subsubsection{\textbf{Feature-based trained metrics}}
Feature-based trained metrics primarily focus on combining various heuristic-based features using a learnable model. These features, obtained from the hypothesis and reference sentences, could be statistical measures such as $n$-gram precision, recall or even untrained metrics such as BLEU or METEOR scores. Further, the learning model can vary from a simple Linear Regressor to a complex Deep Neural Network. We now discuss these different metrics sub-categorized by the learnable model. \\
\\
\textbf{Linear Regression}\\
\textbf{BEER} (BEtter Evaluation as Ranking) \cite{DBLP:conf/wmt/StanojevicS14_beer_wmt,DBLP:conf/emnlp/StanojevicS14_beer}: 
The set of input features used by BEER include precision, recall and F1-score on character $n$-grams for various $n$ and on word-level unigrams. Additionally, they use features based on permutation trees \cite{DBLP:conf/ssst/ZhangG07_pet_permutation_trees} to evaluate word order or fluency. The unigram statistics are computed on function words and content words separately as well as on the entire set of words. 
The BEER model is a simple linear function of the input features given as:
\begin{align*}
    BEER\ score(p,r) = \sum_i W_i x \phi_i(p,r)
\end{align*}
where the different features $\phi_i(p,r)$ are first computed using the hypothesis $p$ and reference sentence $r$, and the model learns the weights $W_i$ for each feature \newA{using linear regression with human judgements from WMT13 \cite{DBLP:conf/wmt/MachacekB13_wmt-13-results} as gold-standard.} \Akash{Are model weights learned from human judgements? Pls write a few sentences on the data and task they used \Ana{Added above, I have tried to verify that other trained metrics also have such details.}}\\
\\
\textbf{SVM Regression}\\
\textbf{BLEND} \cite{DBLP:conf/wmt/MaGWL17_blend}: This metric combines various existing untrained metrics to improve the correlation with human judgements. It uses an SVM regressor with 57 metric scores as features and the DA scores (direct assessment scores on translation quality obtained through human evaluators \cite{DBLP:conf/acllaw/GrahamBMZ13_DA_scores}) from WMT15\cite{DBLP:conf/wmt/StanojevicKKB15_wmt-15-results} and WMT16\cite{DBLP:conf/wmt/BojarGKS16_wmt-16-results} as the gold standard target. The metrics are classified into 3 categories as lexical, syntactic and semantic based metrics. Out of the 57 metrics, 25 are categorized as lexical-based, which correspond only to 9 types of metrics, since some of them are simply different variants of the same metric. \newA{For instance, eight variants of BLEU are formed by using different combinations of $n$-gram lengths, with or without smoothing, etc.} \Akash{can u give an example on what the different variants are for a specific metric say Bleu? \Ana{Added above.}} These 9 metrics are BLEU, NIST, GTM, METEOR, ROUGE, Ol, WER, TER and PER. \newA{17 syntactic metrics are borrowed from the Asiya toolkit \cite{DBLP:journals/pbml/GimenezM10_asiya} along with 13 semantic metrics, which in reality correspond to 3 distinct metrics, related to Named entities, Semantic Roles and Discourse Representation.} \Akash{can u give some examples/insights on these 17 syntactic and 3 semantic metrics? \Ana{None of these details are in the paper. They mainly say they follow DPMFcomb in identifying these 57 metrics. I've checked that and other papers it refers to to add a few insights for these.}} The authors performed an ablation study to analyse the contribution of each of the categories and found that a combination of all the categories provides the best results. \\
\\
\textbf{Grid search with bagging}\\
\textbf{Q-Metrics}: \citet{DBLP:conf/emnlp/NemaK18_qbleu} focus on improving existing $n$-gram metrics such as BLEU, METEOR, ROUGE to obtain a better correlation with human judgements on the answerability criteria for the task of question generation. The authors argue that some words in the hypothesis and reference questions carry more importance that the others and hence propose to assign different weightages to words rather than having equal weights like in standard $n$-gram metrics. 
Hence they categorize the words of the hypothesis and reference question into four categories \textit{viz.} function words, question words (7 Wh-words including `how'), named entities and content words (identified as belonging to none of the previous categories). The $n$-gram precision and recall are computed separately for each of these categories and a weighted average of them to computed to obtain $P_{avg}$ and $R_{avg}$. The Answerability score and Q-metric is defined as:
$$ Answerability = 2.\frac{P_{avg}R_{avg}}{P_{avg} + R_{avg}} $$
$$ Q\text{-}Metric = \delta Answerability + (1-\delta)Metric$$
where $Metric \in \{BLEU, NIST, METEOR, ROUGE\}$. The weights and $\delta$ are tuned using grid search and bagging to find the optimal values that maximize correlation with human scores.\\
\\
\textbf{Neural networks/ Deep Learning}\\
\textbf{Composite metrics:} \citet{DBLP:conf/acl/SharifWBS18_composite_metrics_ic} propose a set of metrics by training a multi-layer feedforward neural network with various combinations of METEOR, CIDEr, WMD, and SPICE metrics as input features. The neural network classifies the hypothesis image caption as either machine-generated or human-generated. The model is trained on Flicker30k \cite{DBLP:conf/iccv/PlummerWCCHL15_flickr30k} dataset, by using 3 out of the 5 reference captions available for each image as positive samples, and captions generated by 3 different models (Show and Tell \cite{DBLP:conf/cvpr/VinyalsTBE15_st_coco}, Show, Attend and Tell \cite{DBLP:conf/icml/XuBKCCSZB15_sat} and Adaptive Attention \cite{DBLP:conf/cvpr/LuXPS17_AA_IC_model}) as negative training samples. \textbf{NNEval} \cite{DBLP:conf/eccv/SharifWBS18_nneval} proposed by the same authors additionally considers BLEU(1-4) scores in the feature set input to the neural network.
\subsubsection{\textbf{End-to-end Trained metrics}}
End-to-end Trained metrics are directly trained using the hypothesis and reference sentences. Note that all the proposed end-to-end trained metrics are based on neural networks. Most of these metrics employ feed-forward neural networks or RNN based models with static/contextualized word embeddings. However, recently pretrained transformer models are also being used in a few metrics.\\
\\
\textbf{SIMILE:} To facilitate better comparison of hypothesis and reference sentences, \citet{DBLP:conf/acl/WietingBGN19_beyondbleu} train a sentence encoder, $g$, on a set of paraphrase pairs (from ParaNMT corpus \cite{wieting-gimpel-2018-paranmt}) using the max margin loss:
$$ l(s,s') = max\Big(0,\delta - cos(g(s),g(s')) + cos(g(s),g(t)) \Big)$$ where $\delta$ is the margin, $s$ and $s'$ are paraphrases, and $t$ is a negative example obtained by random sampling the other sentence pairs.\\
The authors define the metric `SIM' as the cosine similarity of the sentence embeddings of the reference and candidate sentences. To discourage model generations that have repeating words with longer lengths, a length penalty (LP) term is employed in contrast to the Brevity Penalty term in BLEU.
$$ LP(r,p) = e^{1-\frac{max(|r|,|p|)}{min(|r|,|p|)}}$$
Finally SIMILE is defined using SIM and LP as: 
\begin{align*}
    SIMILE = LP(r,p)^\alpha SIM(r,p)
\end{align*}
where $\alpha$ determines the influence of the length penalty term and is tuned over the set {0.25,0.5}.\\
\\
\textbf{ESIM} (Enhanced Sequential Inference Model): ESIM is a model for natural language inference proposed by \citet{DBLP:conf/acl/ChenZLWJI17_ESIM_for_inference}, which has been directly adopted for the task of translation evaluation by \citet{DBLP:conf/acl/MathurBC19_BERTr_contextualized_embeddings_translation}. It consists of a trained BiLSTM model to first compute sentence representations of the reference and hypothesis. Next, the similarity between the reference and hypothesis is calculated using a cross-sentence attention mechanism. These attention weighted representations are then combined to generate \textit{enhanced} representations of the hypothesis and the reference. The enhanced representations are passed as input to another BiLSTM. The max-pooled and average-pooled hidden states of the final BiLSTM are used to predict the ESIM score:
\begin{gather*}
    x = [v_{r,avg};v_{r,max};v_{p,avg};v_{p,max}]
    \\
    ESIM = U^T ReLU(W^Tx+b) +b'
\end{gather*}
where $v_{s,avg/max}$ denotes the average or max pooled vector of the final BiLSTM hidden states for sentence $s$, and $U, W, b and b'$ are parameters to be learnt. 
The metric is trained on the \textit{Direct Assessment} human evaluation data that is collected for WMT 2016 \cite{DBLP:conf/wmt/BojarCFGHHJKLMN16_wmt16}.\\
\\
\textbf{RUSE} (Regressor Using Sentence Enbeddings \cite{DBLP:conf/wmt/ShimanakaKK18_ruse}): RUSE is a MultiLayer Perceptron (MLP) based regression model that combines three pre-trained sentence embeddings. The three types of sentence embeddings used are InferSent \cite{conneau-etal-2017-supervised_InferSent}, Quick-Thought \cite{DBLP:conf/iclr/LogeswaranL18_QuickThought} and Universal Sentence Encoder\cite{DBLP:journals/corr/abs-1803-11175_usn}. Through these embeddings, RUSE aims to utilize the global sentence information that cannot be captured by any local features that are based on character or word $n$-grams. An MLP regressor predicts the RUSE score by using a combination of the sentence embeddings of the hypothesis and reference.
\begin{gather*}
    \overrightarrow{s} = \text{Encoder(s) = [InferSent(s); Quick-Thought(s); UniversalSentenceEncoder(s)]}
    \\
    \text{RUSE = MLP-Regressor}\big( \overrightarrow{p};\overrightarrow{r};|\overrightarrow{p}-\overrightarrow{r}|;\overrightarrow{p}*\overrightarrow{r} ) \big)
\end{gather*}
The sentence embeddings are obtained from pre-trained models and only the MLP regressor is trained on human judgements from the WMT shared tasks over the years 2015-2017.\\ 
\\
\textbf{Transformer based trained metrics}\\
Transformer architecture \cite{DBLP:conf/nips/VaswaniSPUJGKP17_attention_is_all_you_need_transformers} eschews the well-established route of using Recurrent Neural Networks (RNNs and any of its variants) for tasks in NLP (Natural Language Processing). It instead incorporates multiple levels of feed-forward neural networks with attention components. The transformer-based models such as BERT\cite{DBLP:conf/naacl/DevlinCLT19bert}, RoBERTa\cite{DBLP:journals/corr/abs-1907-11692_roberta}, XLNet\cite{DBLP:conf/nips/YangDYCSL19_xlnet}, etc, have shown a lot of promise in various NLP/NLG tasks and have also forayed into the domain of trained evaluation metrics for NLG. We present the transformer-based metrics here.\\
\\
\textbf{BERT for MTE} \cite{shimanaka19} : This model encodes the reference and hypothesis sentences together by concatenating them and passing them through BERT. A `[SEP]' token is added for separation and a `[CLS]' token is prepended to the pair as per the input-requirements of BERT. An MLP-regressor on top of the final representation of the [CLS] token provides the score. Unlike in RUSE, the pretrained BERT encoder is also jointly finetuned for the evaluation task. The other difference from RUSE is the usage of the \textit{pair-encoding} of the candidate and reference sentences together instead of using separate sentence embeddings. The authors report an improvement in correlations with this approach over RUSE.
\begin{gather*}
    \overrightarrow{v} = \text{BERT pair-encoder([CLS] ; p ; [SEP] ; r ; [SEP])}
    \\
    \text{BERT for MTE = MLP-Regressor}\big(\overrightarrow{v}_{[CLS]} \big)
\end{gather*}
\\
\textbf{BLEURT: } \citet{DBLP:journals/corr/abs-2004-04696_bleurt} pretrained BERT with synthetically generated sentence pairs obtained by perturbing Wikipedia sentences via mask-filling with BERT, back-translation or randomly dropping words. A set of pretraining signals are employed including:\\
(i) BLEU, ROUGE and BERTscore, (the latter 2 are split into 3 signals each, using the precision, recall and F-score),\\
(ii) back-translation likelihood indicating the probability that the two sentences are back-translations of the other with either German or French as the intermediate language,\\
(iii) textual entailment signal (indicating Entailment, Contradiction or Neutral) obtained from BERT fine-tuned on entailment task with MNLI (Multi-Genre Natural Language Inference) dataset\cite{williams-etal-2018-broad_mnli}.\\
(iv) back-translation flag to indicate if the perturbation was actually generated through back-translation or mask-filling.\\
The various signal/task-level losses are aggregated using weighted sum. The BLEURT rating is obtained using a linear layer on the embedding produced for the prepended [CLS] token:
$$ BLUERT\ score = \hat{y} = f(r,p) = W\tilde{v}_{[cls]}+b $$
The original BERT, further pretrained on synthetic data with the pretraining signals is finetuned on the task-specific supervised data using regression loss.
$$ loss= \frac{1}{N}\sum_{n=1}^N||y_i-\hat{y}||^2$$
BLEURT achieves state-of-the-art performance on WMT and WebNLG challenges after finetuning on those datasets.\\
\\
\textbf{NUBIA:} \citet{DBLP:journals/corr/abs-2004-14667_nubia} propose a 3-stage architecture called NUBIA for NLG evaluation. The first step is neural feature extraction using various transformer based architectures to represent sentence similarity, logical inference and sentence likelihood/legibility.
The models used for this are:
\begin{itemize}
    \item RoBERTa large pretrained model, finetuned on STS-B-benchmark dataset to predict sentence similarity between hypothesis and reference 
    \item RoBERTa large pretrained model, finetuned on MNLI challenge of GLUE for capturing logical relationship between the hypothesis and reference.
    \item GPT-2 model's perplexity score to determine the grammatical correctness of the hypothesis.
\end{itemize}
The next step/module is termed an aggregator which is either a linear regression model or a feed forward neural network trained to provide a quality score on the hypothesis on its interchangeability with the reference. Finally the calibration step ensures the final value is between 0 and 1, and also that providing the reference sentence as the hypothesis generates a score of 1. The authors show NUBIA outperforms/matches the metrics used to evaluate machine translation, and image captioning in terms of correlations with human judgements.


\section{Context-dependent metrics}
\label{sec:context_dependent_metrics}
In this section, we describe the context-dependent metrics which also consider the input context while evaluating a hypothesis. Similar to the context-free metrics, we categorize the context-dependent metrics into (i) Untrained, and (ii) Trained metrics. We discuss the metrics in these two categories in the next two subsections. We wish to note that since the input context varies across the tasks, context-dependent metrics are specific to their corresponding tasks and cannot be used as it is for other tasks. This is in contrast to context-free metrics like BLEU, ROUGE which are adopted for a wide range of tasks. 

\subsection{Untrained metrics}
Untrained context-dependent metrics can be further classified into two classes based on the type of features they use \textit{viz.} (i) word-based, and (ii) embedding-based. We discuss these in detail below:

\subsubsection{\textbf{Word based metrics}}
Word based context-free metrics evaluate a hypothesis by using the word or $n$-gram features of the hypothesis and the context. ROUGE-C \cite{DBLP:conf/grc/HeCMGLSW08_rouge-c_with_context} and PARENT \cite{DBLP:conf/acl/DhingraFPCDC19_parent} are two metrics under this category. 

\hfill\\
\textbf{ROUGE-C: } In the context of abstractive summarization, \citet{DBLP:conf/grc/HeCMGLSW08_rouge-c_with_context} proposed a modification to ROUGE, dubbed ROUGE-C, where the candidate summary is compared with the document to be summarized, instead of the reference summary. For instance, ROUGE-C-N is given as:
\begin{align*}
     \text{ROUGE-C-N}  = \frac{\sum\limits_{s_h \in \text{hypothesis}} \sum\limits_{\text{$n$-gram} \in s_h} Count_{match}(\text{$n$-gram})}{\sum\limits_{s_c \in \text{Source Document}} \sum\limits_{\text{$n$-gram} \in s_c} Count(\text{$n$-gram})}
\end{align*}
where $s_h$ and $s_c$ are sentences belonging to the summary and the document respectively. ROUGE-C is especially beneficial in cases where a reference summary is not available. Additionally for query-focused summarization task, \textit{i.e.}, the task of creating a summary that answers the given query from the document, the ROUGE-C score is computed as:
\begin{align*}
    \text{ROUGE-C} = \lambda\cdot\text{ROUGE-C}_{QF} + (1-\lambda)\cdot\text{ROUGE-C}_{D}
\end{align*}
where ROUGE-C$_{D}$ is the ROUGE-C score when the document is used as the context and ROUGE-C$_{QF}$ is the ROUGE-C score when the query-focused information, such as the questions, viewpoints, task descriptions, \textit{etc.}, are used as the context. \newA{The weighting factor $\lambda$ is varied from 0 to 1 to check for Pearson/Spearman's correlations with human judgements and original ROUGE scores on DUC (Document Understanding Conference) data\footnote{https://duc.nist.gov/}. }\\
\MK{Write a sentence about $\lambda$. How is it tuned? \Ana{Added some details. They don't explicitly specify what value of $\lambda$ they use. They just show a plot varying $\lambda$ vs correlations with human scores as well as with the original ROUGE-variants scores and make a comment that no one value is the best for all these correlations.}}\\
\\
\textbf{PARENT} (Precision And Recall of Entailed N-grams from the Table): \citet{DBLP:conf/acl/DhingraFPCDC19_parent} proposed the PARENT evaluation metric for the the task of data-to-text generation. PARENT matches the $n$-grams in the hypothesis with both the reference as well as the record/tuple $t$. In order to match the semi-structured data in the table with the unstructured hypothesis, an entailment probability is defined as the probability of an $n$-gram being correct/valid, given the table. The entailment probability for an $n$-gram $g$ is computed either using a word-overlap model, that computes the fraction of words in the $n$-gram that are in the table $t$, \textit{i.e.}, $Pr(g)=\sum_{w\in g}\mathbbm{1}(w\in t)/n$, or using a co-occurrence model that first learns the probability of entailment of each word $Pr(w)$ using co-occurrence counts of words from a training set of table-reference pairs. The entailment probability of the $n$-gram $g$ is then given as the geometric mean of the entailment probabilities of the constituent words. $Pr(g) = \big(\prod_{w\in g} Pr(w)\big)^{1/n}$. 

Entailed precision $P^{Ent}_n$ and entailed recall $R^{Ent}_{n}$ are computed by giving each $n$-gram $g$ a reward of 1 if it overlaps with the reference and a reward proportional to its table entailment probability otherwise. Formally, the entailed precision is given as:
\begin{align*}
    P^{Ent}_n = \frac{\sum_{g\in p} [Pr(g) + (1-Pr(g)).\mathbbm{1}(g \in r) ]}{\sum_{g\in p}1}
\end{align*}
Entailed recall computes recall against the reference $R^{Ent}_{n}(r)$ and the table $R^{Ent}_{n}(t)$ separately and considers the weighted geometric average of them (with a $\lambda$ weight parameter) as follows:
\begin{align*}
    R^{Ent}_{n} = R^{Ent}_{n}(r)^{(1-\lambda)}.R^{Ent}_{n}(t)^\lambda
\end{align*}
The $P^{Ent}_n$ and $R^{Ent}_{n}$ for various $n$'s are aggregated using the geometric mean to get the combined precision $P^{Ent}$ and recall $R^{Ent}$. Finally the PARENT score for each instance is the F-score of the combined precision and recall scores. 
\begin{align*}
    PARENT = \frac{2P^{Ent}R^{Ent}}{P^{Ent} + R^{Ent}} 
\end{align*}
To compare with corpus-level metrics \textit{such as BLEU}, the corpus-level PARENT score is given as the average of instance level PARENT scores. 
\subsubsection{\textbf{Embedding based metrics}}
\hfill\\
\textbf{YiSi-2 }by \citet{DBLP:conf/wmt/Lo19_yisi} as well, is the same as YiSi-1, except that it uses cross-lingual embeddings to compute the similarity of the MT output with the source. That is, YiSi-2 is the bilingual, reference-less version for MT quality estimation, which uses the contextual embeddings extracted from multi-lingual BERT to evaluate the cross lingual lexical semantic similarity between the input and MT output. It can optionally use the shallow semantic parsing module.\\

\subsection{Trained metrics}
In this section, we describe the context-free metrics that contain learnable components which are trained for automatic evaluation. All the proposed metrics in this category are Neural Network based with different architectures ranging from MultiLayer Perceptrons (MLP) to Transformers. Most of the proposed metrics in this category, such as ADEM, RUBER, MaUde, RoBERTa-evaluator and SSREM are for the task of dialogue evaluation. We also discuss the LEIC metric for the image captioning task. 
\\
\subsubsection{\textbf{End-to-end Trained metrics}}
\hfill\\
\textbf{LEIC} \footnote{There was no explicit name provided for this metric by the authors. The `LEIC' acronym has been adopted from the paper's title, `Learning to Evaluate Image Captioning' by many later works that refer to this model including \cite{bertscore,DBLP:conf/emnlp/ZhaoPLGME19_moverscore}}: \citet{DBLP:conf/cvpr/CuiYVHB18_discriminate_ic_eval} observe that the commonly adopted metrics for image captioning evaluation such as CIDEr, METEOR, ROUGE and BLEU mainly focus on the word-overlap between the hypothesis and reference captions and do not correlate well with human judgements. Although SPICE constructs scene graphs from the hypothesis and reference in order to compare semantic similarity, it fails to capture the syntactic structure or fluency of a sentence. For instance, it can be gamed with repetitive sentences as shown in \cite{DBLP:conf/iccv/LiuZYG017_spider}. Moreover, all these rule-based metrics rely solely on similarity between candidate and reference captions, ignoring the image. LEIC is a discriminative evaluation metric that is trained to distinguish between the human and machine-generated captions, by taking the image into account. The image is encoded using ResNet pretrained on ImageNet with fixed weights, and the candidate as well as the reference captions are encoded using an LSTM-based sentence encoder. These encoded feature vectors are combined into a single vector in two different ways which were found to yield comparable results empirically. The first method uses a concatenation of all the vectors followed by a MultiLayer Perceptron (MLP):
\begin{align*}
    v = ReLU(W\cdot \text{concat}([i;r;h]) + b)
\end{align*}
where ReLU (Rectified Linear Units) is an activation function given as $ReLU(x) = max(x,0)$, and $W,b$ are parameters. 
The second method is to concatenate the image and reference first, and then combine it with the candidate caption using `Compact Bilinear Pooling' \cite{DBLP:conf/cvpr/GaoBZD16_cbp}, which has been shown to effectively combine the heterogeneous information of image and text \cite{DBLP:conf/emnlp/FukuiPYRDR16_cbp_effect}. 
The final classifier is trained on the combined feature vectors using a softmax classifier with the cross-entropy loss function. COCO dataset \cite{DBLP:conf/cvpr/VinyalsTBE15_st_coco} is used for training and machine generated captions are obtained using 3 image captioning models proposed in \cite{DBLP:conf/cvpr/KarpathyL15_nt,DBLP:conf/cvpr/VinyalsTBE15_st_coco,DBLP:conf/icml/XuBKCCSZB15_sat}. Further, in order to enable the model to identify pathological captions, data augmentation is performed by (i) randomly sampling captions, (ii) permuting the caption word order and (iii) randomly replacing words in the captions. These are explicitly added as negative examples during training.\\
\\
\textbf{ADEM} (Automatic Dialogue Evaluation Model): ADEM \cite{DBLP:conf/acl/LoweNSABP17ADEM} is trained to evaluate responses, given the dialogue context and a reference response, on a scale of 1 to 5. For this, the authors collect human ratings on a variety of responses for contexts from the Twitter corpus \cite{DBLP:conf/naacl/RitterCD10Twit}. These responses are obtained from dialogue models \cite{lowe-etal-2015-ubuntu,HRED} as well as humans, and thus contain a mix of good and bad responses. This forms the training data where each instance contains the context, the reference response, the proposed response (which could be human generated or system generated) and the score assigned to the proposed response by human evaluators.
ADEM first encodes the context, proposed response and reference response using a Hierarchical-RNN \cite{hrnn} encoder consisting of utterance-level and context-level RNNs. These vector representations of the dialogue context $\overrightarrow{c}$, reference response $\overrightarrow{r}$, and proposed response $\overrightarrow{p}$ are used to compute the evaluation score as follows:
\begin{equation*}
    \mbox{ADEM\ Score}(c,r,p) = (\overrightarrow{c}^T .M. \overrightarrow{p} + \overrightarrow{r}^T. N. \overrightarrow{p} - \alpha)/ \beta
\end{equation*}
where $M$, $N \in \mathbb{R}^{n \times n}$ are learned matrices, and $\alpha, \beta$ are scalar constants used to re-scale scores in the closed range $[1, 5]$. 
The model is trained to minimize the squared error between the model predictions and the human scores with L2-regularization.
\begin{align*}
	\mathcal{L}=\sum_{i=1:K}[score(c_i,r_i,p_i)-human_i]^2+\gamma||\theta||_2
\end{align*}
\\
\textbf{RUBER} (Referenced metric and Unreferenced metric Blended Evaluation Routine \cite{DBLP:conf/aaai/TaoMZY18ruber}): RUBER is another RNN-based dialogue evaluation model like ADEM. However, while ADEM requires human scores on the responses, which are difficult to obtain at large-scale, RUBER proposes a slightly different approach to use unlabelled data. Specifically, RUBER uses a combination of a referenced metric and an unreferenced metric to score the dialogue response on a scale 0 to 1. Here, the term ``referenced'' indicates the need for a reference response whereas ``unreferenced'' means that only the context and the proposed response are considered.


The referenced metric computes the similarity of the reference and proposed response using a modified variant of the vector extrema which is referred to as vector pooling. More specifically, the maximum and minimum values in each dimension are chosen from all the embeddings of the words in a sentence. The closeness of a sentence pair is measured using the cosine similarity of the concatenated max and min-pool vectors.

For the unreferenced metric, a model consisting of bidirectional gated recurrent units (GRUs)\cite{gru} as encoders is learnt. The context and proposed response are passed through the GRUs and the last states of both directions are concatenated to obtain the context and response embeddings $\overrightarrow{c}$ and $\overrightarrow{p}$. These embeddings are concatenated along with a `quadratic feature' defined as $\overrightarrow{c}^TM\overrightarrow{p}$, where $M$ is a parameter matrix. Finally a multi-layer perceptron is used to predict a scalar score that measures the relatedness between the dialogue context and a given response. 
\begin{align*}
    s_U(c,p) = MLP(\overrightarrow{c};\ \overrightarrow{c}^TM\overrightarrow{p};\ \overrightarrow{p})
\end{align*}

To train the unreferenced metric, the authors adopt negative sampling strategy (that is, select random responses belonging to other contexts) to get negative examples or invalid responses for a given context. The model is trained on Chinese dialogue data with the Hinge loss function given by: 
\begin{align*}
    J = \max \{0, \Delta - s_U(c,p^+) + s_U(c,p^-) \}
\end{align*}
where $\Delta$ is a threshold parameter that recommends the score of a positive response $p^+$ to be larger than that of a random negative response $p^-$ by at least a margin $\Delta$.

Finally to get the RUBER score, the referenced and unreferenced metric scores are first normalized to be in the range [0,1] and then combined using an aggregate function such as min, max, geometric averaging or arithmetic averaging. The authors show that all the different aggregate functions produce similar results.\\
\\
\textbf{SSREM} (Speaker Sensitive Response Evaluation Model \cite{DBLP:journals/corr/abs-2006-07015_ssrem} ): The SSREM metric follows the strategy of RUBER to combine a reference-based score and an unreferenced score. SSREM formulation can be compactly written as:
\begin{align*}
    SSREM(c,r,p) = h(f(c,p),g(r,p))
\end{align*}
where $f(c,p) = tanh(\overrightarrow{c}^TM\overrightarrow{p})$ is trained to score the relevance of response $p$ to context $c$ with a parameter matrix $M$. $\overrightarrow{c}$ and $\overrightarrow{p}$ are obtained using vector averaging on the GloVe embeddings \cite{DBLP:conf/emnlp/PenningtonSM14_glove} of the words in the context and response respectively. The function $g(r,p)$ measures the similarity between the proposed response $p$ and reference response $r$ using sentence mover's similarity \cite{DBLP:conf/acl/ClarkCS19_sms_sentence_mover} with Elmo embeddings \cite{DBLP:conf/naacl/PetersNIGCLZ18_elmo}. The $h$ function combines these two scores using arithmetic averaging.

However, unlike RUBER, SSREM uses `speaker sensitive samples', instead of one positive sample and one random negative response sample for each context. The speaker sensitive samples are obtained by considering additional speaker-related information while probing for negative samples, in order to obtain responses that have varying degrees of similarity with the reference response. In decreasing order of similarity as hypothesized by the authors, the samples are obtained using utterances from:\\ (i) the same speaker in the same conversation\\  (ii) the same speaker in a different conversation but with the same partner\\
(iii) the same speaker in a different conversation with a different partner\\
(iv) a random other speaker\\
These cases are empirically verified to be more challenging to distinguish than the randomly sampled negatives. The authors hence train the $f$ function of SSREM to score the reference response higher than all of these negative cases by formulating the loss function as:
\begin{align*}
    -\sum_{c}\log\frac{\exp (f(c,r))}{\sum_{p \in P} \exp(f(c,p))}
\end{align*}
where $P$ is the set of all proposed responses for context $c$ which contains the reference response $r$ along with the negative samples obtained from the above four scenarios ($P$ is thus a set containing the reference response and the four hard negative samples explained above). SSREM is trained using the Twitter corpus since it contains several conversations from each speaker, which is necessary to obtain the speaker sensitive samples. \\
\\
\textbf{Transformer based metrics}\\
\textbf{RUBER with BERT embeddings:} \citet{ghazarian19} propose enhancing RUBER with contextualized embeddings. The modification to the referenced part of RUBER is to simply replace the word2vec embeddings with BERT embeddings. The unreferenced model's architectural changes include (i) replacing the Bi-RNN unit which encodes the word2vec embeddings with a pooling unit which aggregates the BERT embeddings of the words and (ii) replacing the MLP network which outputs a single unreferenced score in RUBER with a binary MLP classifier that assigns label 1 to positive responses and 0 to negative responses.
Unlike RUBER, the authors use the cross entropy loss as the training objective. Note that the BERT embeddings are not finetuned in this case and only the weights of the MLP classifier are learnt by training on the DailyDialog dataset\cite{DBLP:conf/ijcnlp/LiSSLCN17DailyOri}.\\ 
\\
\textbf{MaUde} (Metric for automatic Unreferenced dialogue evaluation) : \citet{DBLP:journals/corr/abs-2005-00583_maude} propose MaUde to score dialogue responses without any need for reference responses. MaUde obtains the BERT encodings for each utterance $u_i$ in the context, which are then downsampled (to reduce the dimensions) using a learnt matrix $D_g$. These downsampled BERT representations $h_{u_i}$ for every utterance $u_i$ are sequentially passed through a bidirectional-LSTM (BiLSTM). The BiLSTM hidden states are combined using max-pooling to get the context representation which is then transformed to the same vector space as the encoded candidate response vector using weight matrix $W$. This transformed context encoding $\overrightarrow{c}$ is combined with the response encoding $\overrightarrow{p}$ and fed to a sigmoid classifier to produce the MaUde score.
\begin{align*}
        \overrightarrow{u_i} &= D_g f_E^{BERT}(u_i)\\
    h_{u_{i+1}}' &= BiLSTM(\overrightarrow{u_i},h_{u_i}')\\
    \overrightarrow{c} &= W.pool_{t \in \{u_1,..,u_{n-1}\}}(h_t')\\
    \overrightarrow{p} &= D_g f_E^{BERT}(p)\\
    score(c, p) &= \sigma(concat([\overrightarrow{p};\overrightarrow{c};\overrightarrow{p}*\overrightarrow{c};\overrightarrow{p}-\overrightarrow{c}]))
\end{align*}
where $f_E^{BERT}$ is the BERT-encoder that is used to encode each utterance, including the candidate response and $h_{u_i}'$ is the hidden representation of $u_i$ in the BiLSTM. The MaUde score is in the range [0,1]. 

MaUde is trained on PersonaChat\cite{personachat}, an open domain chit-chat style dataset. MaUde is trained using two types of negative responses: (i) syntactic negative, and (ii) semantic negatives responses.  Syntactic negative responses are obtained using the following strategies: shuffling the word order, dropping out words randomly, and repeating words in the reference response. Similarly, semantic negatives are obtained by (i) random sampling responses from other dialogues, (ii) using a random response generated for another context by a pretrained seq2seq model on the dataset, and (iii) using a paraphrase of a random negative response, obtained via back-translation \cite{DBLP:conf/emnlp/EdunovOAG18_backtranslation_at_scale}. 
Additionally, a back-translation of the correct (reference) response is used as another positive response.\\
\\
\textbf{RoBERTa-eval:} \citet{DBLP:journals/corr/abs-2004-04908_roberta_eval} utilize the contextualized text embeddings produced by RoBERTa model to encode the dialogue context and proposed response into a single vector $\overrightarrow{d}$. An MLP classifier with sigmoid activation is used to obtain a score between 1 to 5.\\
\begin{align*}
    \overrightarrow{d} &= RoBERTa([c;p];\phi)\\
    \text{RoBERTa-eval}(c,p) &= 4. MLP(\overrightarrow{d};\theta) + 1
\end{align*}
where RoBERTa's parameters $\phi$ can be finetuned and the MLP's parameters $\theta$ are learnt during training. Further, scaling is done to keep the score in the range of 1 to 5. 
ROBERTa-evaluator is trained on DailyDialog dataset \cite{DBLP:conf/ijcnlp/LiSSLCN17DailyOri} using human annotations on response quality. To get a variety of responses of different quality, the authors use random negative sampling, and also obtain responses from generative models proposed in \cite{DBLP:conf/nips/SutskeverVL14_s2s,HRED,vhred,DBLP:journals/corr/abs-1901-08149_transfertransfo}.
\section{Studies criticising the use of automatic evaluation metrics}
\label{sec:criticism}
While automatic metrics have been widely used for evaluating a variety of NLG tasks, there has always been skepticism about their ability to replace human judgements. Indeed, there have been several studies which have criticised the use of such automatic metrics. These studies highlight the (i) poor correlation of these metrics with human judgements (ii) lack of interpretability in these metrics (ii) inherent bias in the metrics (iv) poor adaptability to a wider variety of tasks and (v) inability to capture subtle nuances in language. In this section we briefly summarise such studies which critically examine the use of automatic evaluation metrics for different tasks. Note that most of these studies were done in the context of BLEU and some of the early NLG metrics, such as, NIST, METEOR and ROUGE. This is simply because these metrics have been in use for many years and have been widely adopted even for newer NLG tasks (such as image captioning). As a result of their popularity, they also tend to get scrutinised more critically. \\
\textbf{Poor correlations: }
One recurring criticism of automatic metrics is that they correlate poorly with human judgements. One of the earliest studies in this direction computed the correlation between several automatic metrics and human judgements on fluency and adequacy
\cite{10.1007/978-3-540-30586-6_38_critique1}. They found that BLEU, NIST, SSA (Simple String Accuracy) \cite{DBLP:conf/inlg/BangaloreRW00_ssa}, Melamed's F-measure (i.e., GTM \cite{article_general_text_matcher_gtm}), and LSA (Latent Semantic Analysis)  \cite{DBLP:journals/jasis/DeerwesterDLFH90_LSA} correlate negatively with human judgements on fluency. Further, these metrics showed moderate to less correlation with human adequacy scores. 
Over the years, similar poor correlations with human judgements have been reported for several metrics for various NLG tasks as summarized in Table \ref{tab:critic_corrs}.\\
\begin{table}[h]
    \centering
    \begin{tabular}{C{0.7cm}|L{6.68cm}|c|L{5.58cm}|}
         Work & Metrics & Task & Datasets on which poor correlation was observed\\ \hline
         \cite{DBLP:conf/emnlp/NovikovaDCR17} & BLEU, TER, ROUGE, NIST, LEPOR, METEOR, CIDEr & DG & BAGEL \cite{mairesse-etal-2010_bagel_dataset}, SF-HOTEL, SF-REST \cite{wen-etal-2015_sfhotel_rest_dataset}\\
         \cite{DBLP:conf/acl/ElliottK14} & Smoothed-BLEU \cite{DBLP:conf/acl/ClarkDLS11_smoothedbleu}, TER, ROUGE-SU4, Meteor  & IC & Flickr8K\cite{DBLP:journals/jair/HodoshYH13_flickr8k}, E\& K\cite{DBLP:conf/emnlp/ElliottK13_EKdataset}\\
         \cite{kryscinski-etal-2019-neural_summ_critic} & ROUGE-1,2,L & AS & CNN/DailyMail\cite{DBLP:conf/conll/NallapatiZSGX16_cnn_dailymail_dataset2} \\
         \cite{DBLP:journals/csl/DusekNR20_e2enlg} & BLEU, NIST, METEOR, ROUGE-L, CIDEr & D2T & E2E NLG dataset \cite{DBLP:journals/csl/DusekNR20_e2enlg}\\
         \cite{DBLP:conf/emnlp/NemaK18_qbleu} & BLEU, ROUGE-L, NIST, METEOR& QG & SQuAD\cite{DBLP:conf/emnlp/RajpurkarZLL16_squad_dataset}, WikiMovies\cite{DBLP:conf/emnlp/MillerFDKBW16_wikimovies_dataset}, VQA\cite{DBLP:conf/iccv/AntolALMBZP15_vqa_dataset}\\
         \cite{DBLP:conf/acl/DhingraFPCDC19_parent} & ROUGE, CIDEr, METEOR, BLEU, CS & D2T & WikiBio\cite{DBLP:journals/corr/abs-1910-08684_wikibio_confidence} \\
         \cite{DBLP:journals/corr/abs-1807-02202_chaganty} & BLEU, ROUGE, METEOR, VecSim &QA/AS& MS MARCO\cite{DBLP:conf/nips/NguyenRSGTMD16_msmarco_dataset}, CNN/DailyMail\cite{DBLP:conf/nips/HermannKGEKSB15_cnn_dailymail_dataset1,DBLP:conf/conll/NallapatiZSGX16_cnn_dailymail_dataset2}  \\
         \cite{DBLP:conf/acl-mrqa/ChenSSG19} & BLEU, ROUGE, SMS, BERTScore & QA& NarrativeQA\cite{DBLP:journals/tacl/KociskySBDHMG18_narrativeQA}, SemEval\cite{DBLP:conf/semeval/OstermannRMTP18_semeval_2018_task11}, ROPES\cite{DBLP:conf/acl-mrqa/LinTCG19_ropes_dataset}\\
         \cite{DBLP:conf/emnlp/LiuLSNCP16HowNot} & BLEU, ROUGE, METEOR, Greedy matching, vector extrema, vector avg & DG & Twitter\cite{DBLP:conf/naacl/RitterCD10Twit}, Ubuntu\cite{lowe-etal-2015-ubuntu}\\
    \end{tabular}
    \caption{Works showing poor correlation of various metrics with human judgements on various datasets \MK{should we add one more column on task \Ana{Done}}}
    \label{tab:critic_corrs}
\end{table}\\

In the context of Table \ref{tab:critic_corrs}, it is worth mentioning that there is a high variance in the correlations reported for the same metric across different studies (as also observed in \cite{DBLP:journals/coling/Reiter18_validity_of_bleu}). 
This could be due to differences in the procedures followed to procure human ratings or metric-specific parameter settings used in different studies (which are not always thoroughly reported \cite{DBLP:conf/wmt/Post18_bleu_std_sacrebleu}). To alleviate the influence of such external factors, the WMT shared metrics task, which annually evaluates the metrics used for MT, standardizes the dataset and human evaluation setup. However, despite such careful standardisation there could still be issues. For example, in the context of the recent WMT-19 task, \citet{DBLP:conf/wmt/MaWBG19} study the 24 metrics that were submitted and 
show that correlations reported are not reliable if all the translation systems to be ranked are good systems. 
In \citet{DBLP:journals/corr/abs-2006-06264_tangled_up_in_bleu}, the authors further show that this is a generic problem encountered in any scenario which involves ranking a small set of systems that are all of similar capability.
Also, at the sentence level, \citet{DBLP:journals/coling/FomichevaS19_correlation_1} find that automatic evaluation metrics used for MT evaluation are better at distinguishing between high-quality translations compared to low-quality translations.
Overall, many studies  \cite{DBLP:journals/coling/Reiter18_validity_of_bleu,DBLP:conf/emnlp/NovikovaDCR17,DBLP:conf/acl/DhingraFPCDC19_parent} re-confirm the findings that automatic metrics are reliable (with better correlations and less variations across studies) 
at the system-level \cite{DBLP:journals/coling/ReiterB09} and less so at the sentence-level \cite{10.1007/978-3-540-30586-6_38_critique1}. 
\\
\\
\textbf{Uninterpretability of scores: }
As discussed in section \ref{sec:human_eval}, human evaluators are required to assign multiple scores to a given hypothesis where each score corresponds to a specific criteria (fluency, adequacy, coherence, relevance, thoroughness, \textit{etc.}). However, automatic evaluation metrics assign a single score to a given hypothesis and it is often not clear which of the relevant criteria this score captures or corresponds to. Hence, these scores assigned by automatic metrics are difficult to interpret. For example, in the context of a summarization system, if a metric assigns a low score to a generated output then it is not clear whether this low score corresponds to poor fluency or poor informativeness or poor coherence.  
It is not surprising that within just 2 years from when BLEU and NIST were proposed, these metrics were criticized for their poor interpretability by \citet{DBLP:conf/lrec/ZhangVW04_bleu_nist_quant_imprvment}. 
\citet{callison-burch-etal-2006-evaluating_bleu_in_mt} further demonstrate that an improvement in BLEU score is neither sufficient nor necessary to indicate progress in translation task thereby raising questions about what the score really stands for. In particular, they show BLEU score can be misleading since several permutations of the n-grams of a sentence would get the same score as the original sentence, even though not all of the permutations would be correct or sensible. 
In other words, BLEU admits many spurious variants. It also penalizes correct translations if they substantially differ from the vocabulary of the references. Even the more recent evaluation metrics are uninterpretable as they just assign one score to the hypothesis as opposed to the different criteria used by humans for evaluating such systems.

~\\
\textbf{Bias in the metrics: }
Some studies have also shown biases in specific metrics. For example, BLEU is found to be favourably biased towards n-gram-based translation systems as opposed to rule-based systems \cite{coughlin2003correlating,callison-burch-etal-2006-evaluating_bleu_in_mt}.
In the context of more modern evaluation systems, it was found that GAN-based evaluators \cite{li2017adversarial} have poor generalization. In particular, they are not good at evaluating systems different from the ones that they have been trained upon. 
Similarly, \citet{DBLP:conf/aaai/SaiGKS19Reeval} found that ADEM, which is used for evaluating dialogue responses, always assigns scores in a narrow range around the median, irrespective of whether the generated response is relevant or not. They further observe that the response encodings generated by ADEM are very close to each in the vector space with a very high conicity (\textit{i.e.}, there is no clear separation between the encodings of relevant and irrelevant responses). So far such studies which reveal specific biases in a metric are limited to a few metrics as listed above and it would be interesting to check for some biases in the newer metrics proposed in the recent years. 
\\
\\
\textbf{Poor adaptability across Tasks: }
As mentioned multiple times before, the criteria used for evaluating NLG systems vary across different tasks. As a result, the adoption of a metric proposed for one task for another task is not always prudent and has been criticized in many studies \cite{DBLP:journals/coling/Reiter18_validity_of_bleu,DBLP:conf/emnlp/NemaK18_qbleu,DBLP:conf/emnlp/LiuLSNCP16HowNot,DBLP:conf/eacl/KilickayaEIE17_wmd_ic}. A case in point is the poor choice of using n-gram based metrics such as BLEU, METOER, \textit{etc.}, for evaluating dialog systems\cite{DBLP:conf/emnlp/LiuLSNCP16HowNot}. These metrics check for n-gram based overlap between the reference and the hypothesis, which does not make sense in the context of a dialog system where widely varying responses are possible. In particular, a hypothesis may not have any n-gram overlap with the reference but still be a correct response for the given context.
\hfill \\\\
\textbf{Inability to capture all nuances in a task: }
Even task-specific metrics are unable to account for all the nuances of the task. For example, \citet{kryscinski-etal-2019-neural_summ_critic} criticize the automatic metrics and human evaluations used for abstractive summarization stating that none of them check for factual inconsistencies in the summaries. Similarly, \citet{DBLP:conf/emnlp/WisemanSR17} discuss the lack of a reliable measurement of faithfulness in the context of Data-to-Text Generation. Even dialog specific metrics such as ADEM fail to account for the diversity in valid responses \cite{DBLP:conf/emnlp/LiuLSNCP16HowNot,DBLP:conf/aaai/SaiGKS19Reeval}. 
Similarly, \citet{Ananthakrishnan2006SomeII_more_blues} analyze the Hindi-English translation task and list the various issues and divergence in the language-pair that are not effectively captured by BLEU. These include lexical, categorical, pleonastic, and stylistic divergence. 



\section{Evaluating Evaluation Metrics}
\label{sec:evaluating_metrics}
In this section, we discuss the various methodologies used to assess the effectiveness of automatic evaluation metrics. One of our primary goals here is to identify how well automatic evaluation metrics can act as a proxy for human evaluations. To do so, the most widely used method is to compute the correlation between the scores given by the evaluation metric and human judgements on system generated or synthetically crafted outputs based on several task specific criteria as discussed in section \ref{sec:human_eval}. In the next subsection, we discuss the various commonly used correlation measures and highlight their main differences with examples. 

\subsection{Correlation Measures}
\noindent \textbf{Pearson correlation coefficient} measures the linear dependence between two continuous variables. Pearson correlation assumes that (i) there is a linear relationship between the two variables, and (ii) the two variables follow a bi-variate normal distribution. Given paired data $\{(x_{1},y_{1}),\dots ,(x_{n},y_{n})\}$, the Pearson correlation coefficient $\rho_{xy}$ is defined as follows:

\begin{equation*}
    {\rho_{xy}={\frac {\sum _{i=1}^{n}(x_{i}-{\bar {x}})(y_{i}-{\bar {y}})}{{\sqrt {\sum _{i=1}^{n}(x_{i}-{\bar {x}})^{2}}}{\sqrt {\sum _{i=1}^{n}(y_{i}-{\bar {y}})^{2}}}}}}
\end{equation*}

The value of the Pearson correlation coefficient ranges from -1 to +1. The correlation coefficient is +1 in the case of a perfect increasing linear relationship, -1 in the case of a perfect decreasing linear relationship. If the variables are independent, Pearson's correlation coefficient is 0, but the converse is not true since Pearson correlation only measures the linear dependence between two variables.

There can be situations where human judgements are obtained as a binary variable (good or bad), but the evaluation metric is continuous. In such cases, where one variable is dichotomous and other variable is continuous, the Point biserial correlation 
coefficient is usually used as the correlation measure. Point biserial correlation coefficient is mathematically equivalent to computing the Pearson correlation after assigning two distinct numerical values to the dichotomous variable. \\

\noindent \textbf{Spearman's correlation coefficient} measures the monotonic relationship (whether linear or not) between two variables. The Spearman correlation between two variables is equal to the Pearson correlation between the rank values of those two variables. Mathematically, given paired data $\{(x_{1},y_{1}),\dots ,(x_{n},y_{n})\}$, the Spearman's correlation coefficient $r_{xy}$ is defined as follows:

\begin{equation*}
    {r_{xy}={\frac {\sum _{i=1}^{n}(r_{x_{i}}-{\bar r_{x}})(r_{y_{i}}-{\bar r_{y}})}{{\sqrt {\sum _{i=1}^{n}(r_{x_{i}}-{\bar r_{x}})^{2}}}{\sqrt {\sum _{i=1}^{n}(r_{y_{i}}-{\bar r_{y}})^{2}}}}}}
\end{equation*}

where $r_{x_{i}}$ and $r_{y_{i}}$ are the rank of $x_{i}$ and $y_{i}$ in $x$ and $y$ respectively. Similar to Pearson, Spearman's correlation coefficient takes values from -1 to +1. A perfect Spearman correlation of +1 or -1 
occurs when there is perfect monotonic relationship between the two variables. \\

\noindent \textbf{Kendall's $\tau$ coefficient} is another rank correlation measure that is computed based on the number of observation pairs that are concordant and discordant. Any pair of observations $(x_{i},y_{i})$ and $(x_{j},y_{j})$, where $i<j$, are said to be concordant if either $x_{i}>x_{j}$ and $y_{i}>y_{j}$ holds or $x_{i}<x_{j}$ and $y_{i}<y_{j}$ holds; otherwise they are said to be discordant. The Kendall's $\tau$ coefficient is defined as: 

\begin{equation*}
  \tau ={\frac{(\text{number of concordant pairs})-(\text{number of discordant pairs})}{{n \choose 2}}}
\end{equation*}

Similar to Pearson and Spearman's correlation coefficients, Kendall's $\tau$ coefficient also takes values from -1 to +1. The correlation is +1 or -1 when the ranking of the two variables are exactly same or exactly opposite i.e. when there is perfect monotonic relationship between the two variables. \\
\begin{figure}
     \centering
     \begin{subfigure}[b]{0.24\textwidth}
         \centering
         \captionsetup{justification=centering}
         \includegraphics[width=\textwidth]{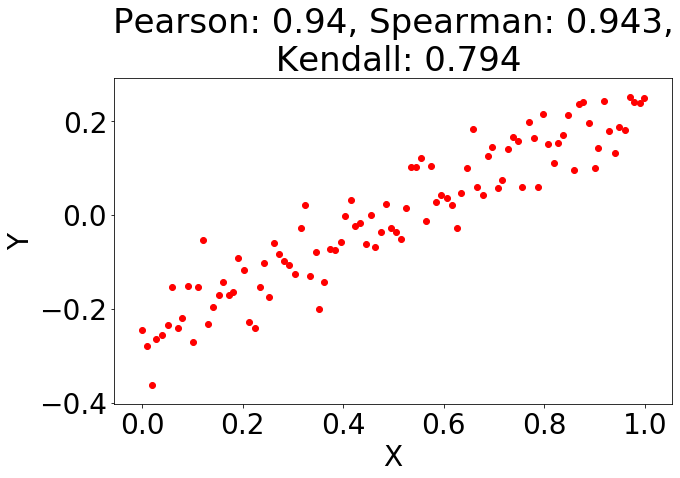}
         \caption{X and Y have a linear relationship. X is uniformly distributed from 0 to 1}
         \label{fig:corr(a)}
     \end{subfigure}
     \hfill
     \begin{subfigure}[b]{0.24\textwidth}
         \centering
        \captionsetup{justification=centering}
         \includegraphics[width=\textwidth]{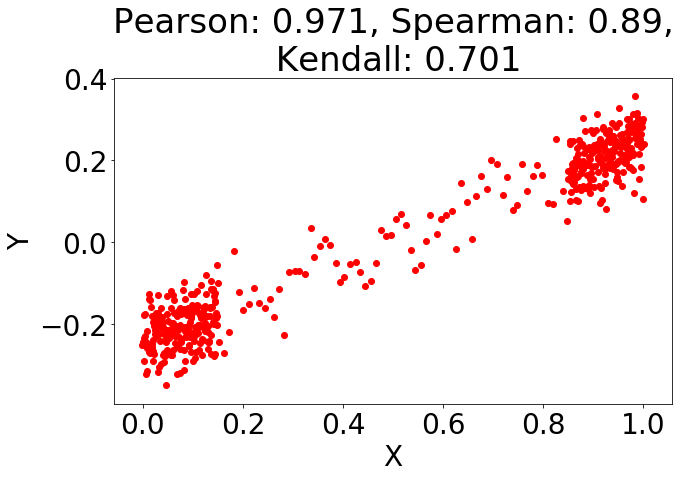}
         \caption{X and Y have a linear relationship. X has a higher density near 0 and 1.}
         \label{fig:corr(b)}
     \end{subfigure}
     \hfill
     \begin{subfigure}[b]{0.24\textwidth}
         \centering
          \captionsetup{justification=centering}
        \includegraphics[width=\textwidth]{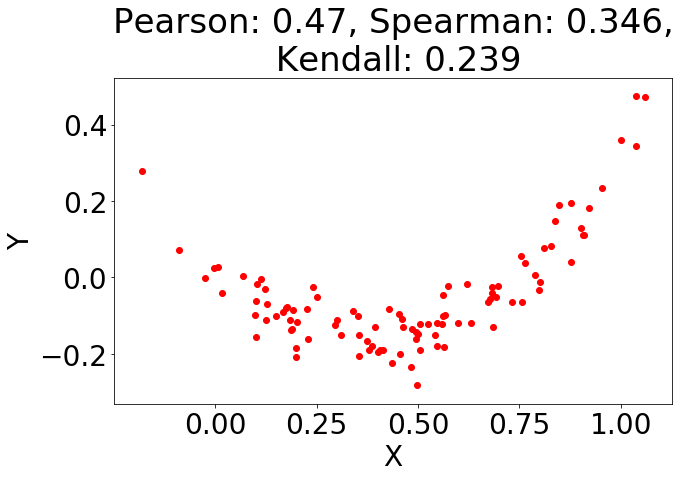}
         \caption{X and Y have a non-monotonic quadratic relationship }
         \label{fig:corr(c)}
     \end{subfigure}
     \hfill
     \begin{subfigure}[b]{0.24\textwidth}
         \centering
          \captionsetup{justification=centering}
        \includegraphics[width=\textwidth]{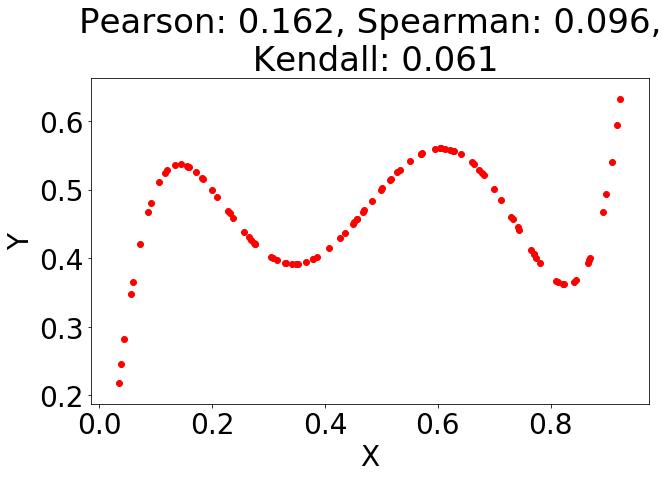}
         \caption{X and Y have a non-monotonic polynomial (of degree 5) relationship }
         \label{fig:corr(d)}
     \end{subfigure}     
    \par\bigskip 
     \centering
     \begin{subfigure}[b]{0.24\textwidth}
         \centering
          \captionsetup{justification=centering}
        \includegraphics[width=\textwidth]{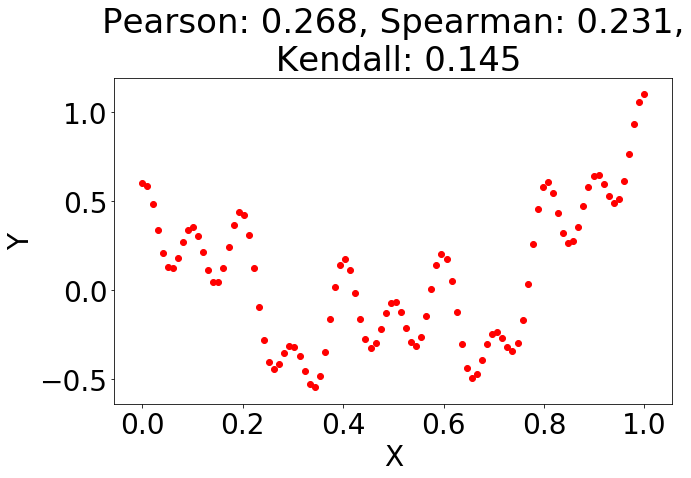}
         \caption{X and Y have an non-monotonic arbitrary relationship}
         \label{fig:corr(e)}
     \end{subfigure}
     \hfill
     \begin{subfigure}[b]{0.24\textwidth}
         \centering
        \captionsetup{justification=centering}
        \includegraphics[width=\textwidth]{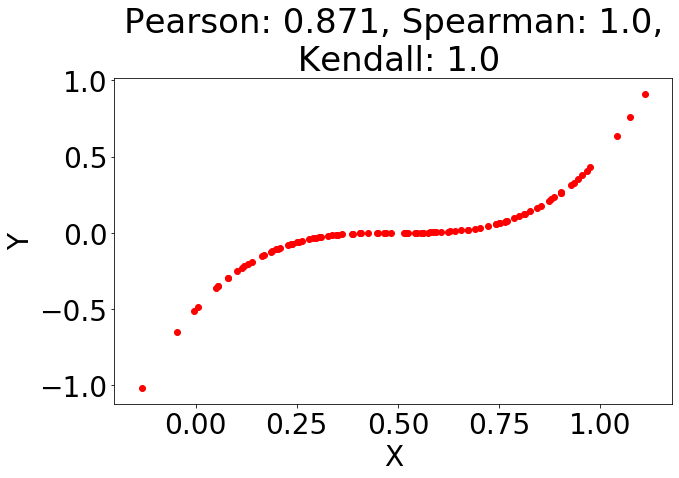}
         \caption{X and Y have a perfect non-linear, monotonic relationship}
         \label{fig:corr(f)}
     \end{subfigure}
     \hfill
     \begin{subfigure}[b]{0.24\textwidth}
         \centering
        \captionsetup{justification=centering}
        \includegraphics[width=\textwidth]{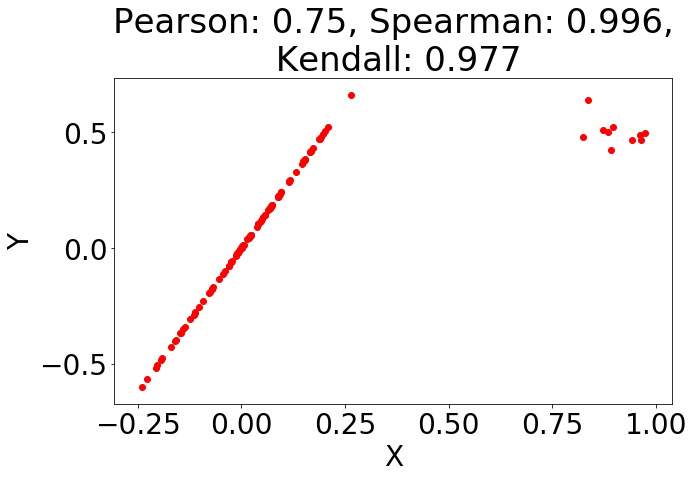}
         \caption{X and Y have a linear relationship with outliers at the tails}
         \label{fig:corr(g)}
     \end{subfigure}
     \hfill
     \begin{subfigure}[b]{0.24\textwidth}
         \centering
        \captionsetup{justification=centering}
         \includegraphics[width=\textwidth]{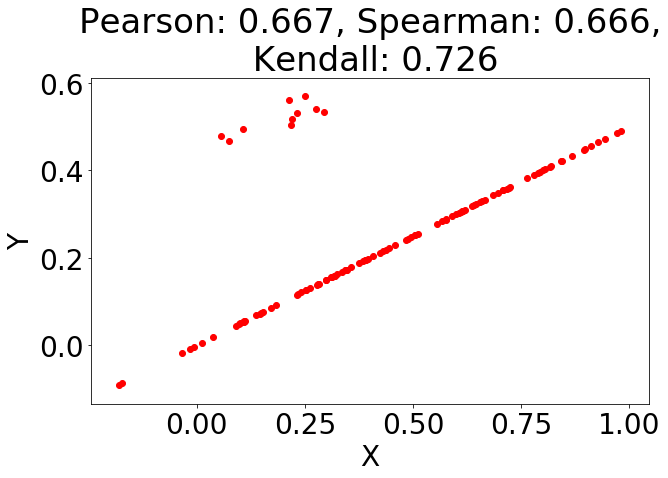}
         \caption{X and Y have a linear relationship with outliers at the middle}
         \label{fig:corr(h)}
     \end{subfigure}
    \caption{Figure illustrating the difference between Pearson, Spearman, Kendall's $\tau$ coefficient under different scenarios.}
    \label{fig:corr}
\end{figure}

\subsubsection{Discussion:}
We shall now discuss the differences between the above discussed correlation measures with help of examples in Figure \ref{fig:corr}. Consider the example in Figure \ref{fig:corr(a)} where the two variables X, Y have a linear relationship. As expected, the Pearson and Spearman's correlations are high. However, due to the presence of noise, many neighbouring observation pairs are discordant, and hence the Kendall's $\tau$ correlation is lower. Furthermore, whenever the values of a variable differ only slightly, rank based metrics can give an unfair penalization since they are only concerned with the rank of the two variables and not their actual values. This trend is observed more clearly in Spearman's correlation in Figure \ref{fig:corr(b)} where there is a higher density of observations near X=0 and X=1. In fact, many NLG system pairs indeed have very small score differences \cite{machacek-bojar-2014-results}, and evaluating with Kendall's $\tau$ or Spearman's correlations harshly penalises metrics that have a different ordering for these systems. For this reason, the WMT shared metric task has shifted to using Pearson correlation at the system-level since 2014 \cite{machacek-bojar-2014-results, DBLP:journals/corr/abs-2006-06264_tangled_up_in_bleu}. However, there are other issues with Pearson correlation. When the relationship between the two variables are non-monotonic (without any outliers) as show in figure \ref{fig:corr(c)}, \ref{fig:corr(d)}, \ref{fig:corr(e)}, all the correlation metrics correctly provides a low score. But as shown in figure \ref{fig:corr(f)}, when the two variables have a perfect monotonic relationship, the Spearman and Kendall's $tau$ have a perfect correlation but the Pearson correlation is lower. Pearson correlation can only reflect whether the relation between the two variables is linear (as opposed to, say, quadratic). Pearson correlation is also highly sensitive to outliers \cite{osborne2004power}. As illustrated in figure \ref{fig:corr(g)}, the Pearson correlation changes drastically in the presence of outliers. Spearman and Kendall's $\tau$ are less sensitive to strong outliers that are in the tails of both samples since the ranking of other points don't change significantly. It's worth noting that Spearman and Kendall's $\tau$ correlation can be sensitive to outliers when the outliers are in the middle as shown in figure \ref{fig:corr(h)}.  Taking these limitations of Pearson and rank based correlations into consideration, the organizers of the WMT Metrics Shared Task \cite{DBLP:conf/wmt/BojarGK17_wmt-17-results,wmt-2018-results,DBLP:conf/wmt/MaWBG19} used a custom variant of Kendall's $\tau$ to account for potential noise in the human ratings. The organisers discard all the translation pairs for the same reference segment which have a “similar” score (less than 25 points away on a 100 points scale). The number of concordant and discordant pairs are then calculated from the remaining pairs, which have a clear difference in the human judgements. \\

\subsection{Significance Tests:}
While the correlation co-efficient quantifies the strength of the dependence between two variables, we do not know whether the correlation is statistically significant or not. For example, a Pearson correlation of 0.7 between human judgements and metric scores from 500 samples would be much more significant than the same correlation from 5 samples. In particular, we could have obtained a correlation of 0.7 from 5 samples just due to chance. Therefore, correlation co-efficients are usually supported by the p-value of a hypothesis test to examine the evidence against the null-hypothesis: the population correlation coefficient is zero. A smaller p-value means that there is stronger evidence in favor of the alternative hypothesis \textit{i.e.} the population correlation is non-zero. The p-value would depend on various factors such as the sample size, the variations in the data, etc. \\

Furthermore, significance tests can be carried out to find 
if the differences in two correlations are statistically significant. For example, consider the case where correlation of two different metrics with 500 human judgements are 0.70 and 0.72. While we can infer whether individual correlations are be statistically significant or not using the p-values, we cannot say whether the difference between the two metrics is significant or not based on these p-values. To determine whether this difference is significant, the widely adopted William's test is used. \\

\noindent \textbf{William's test} \cite{williams-test} evaluates the significance between two dependent correlations sharing one variable. The null hypothesis of the William's test is defined as follows: The population correlation between $X_1$ and $X_3$ equals the population correlation between $X_2$ and $X_3$. In our case, $X_1$ and $X_2$ are the scores of the two metrics and $X_3$ corresponds to human judgement scores. The test statistic for the hypothesis test is:
\begin{align*}
    t(n-3) = \frac{(r_{13}-r_{23}) \sqrt{(n-1)(1+r_{12})}}{\sqrt{2K\frac{(n-1)}{(n-3)} + \frac{{(r_{23}+r_{13})}^{2}}{4} {(1-r_{12})}^3 }}
\end{align*}

where $r_{ij}$ is the Pearson correlation between $X_i$
and $X_j$, $n$ is the size of the population, and: $K = 1 - r_{12}^2 - r_{13}^2 - r_{23}^2 + 2r_{12}r_{13}r_{23}$. 

\subsection{Issues and Influence of the Data and Human-evaluation setup}
So far, in this section, we have mentioned that the most widely used technique for judging the effectiveness of a metric. We now discuss some studies which talk about how the data 
and/or the human evaluation setup can affect the correlation scores. \citet{DBLP:conf/coling/LinO04_orange} were among the earliest to show the shortcomings of correlation based studies in the context of MT, suggesting that human judgements are inconsistent and expensive. Instead they propose to validate how each metric ranks the reference and the n-best machine generated translations. They calculate  the  average  rank  of   the   references in the n-best list, and compute the ratio of  the  average  reference  rank  to  the  length  of the n-best list. This ratio is termed “ORANGE” (Oracle Ranking for Gisting Evaluation) and the smaller the ratio is, the better the automatic metric is.  This strategy works for MT systems which provide n-best translations. \citet{article_early_mt_eval_comp} compare the various automatic metrics and human-based metrics for translation and have demonstrated the effect of the quality of references on automatic metrics, as well as the subjectiveness of the human judges in human-based evaluations.  \citet{belz-reiter-2006-comparing_auto_human_nlg_eval} stress on the influence of corpus-quality on the correlations between human and automatic evaluation schemes, observing that large number of good quality references lead to better correlations. \citet{DBLP:conf/acl/DhingraFPCDC19_parent} show that references aren't always the perfect gold-standard that they're assumed to be by most automatic metrics, especially in the datasets collected automatically and heuristically.
\cite{DBLP:conf/emnlp/NovikovaDCR17} also assess the quality of the data and examine its characteristics such as the length of the references, words,  characters, or syllables per sentence, etc. in the dataset and find that these measures influence the correlations. 
\cite{kryscinski-etal-2019-neural_summ_critic} call attention to potential problems in the datasets including layout bias in news datasets and noise in any web-scraped datasets. \cite{DBLP:journals/corr/abs-2006-06264_tangled_up_in_bleu} observe eliminating outliers (and following other relevant bias and noise eliminating strategies) can lead to more reliable and robust correlations to compare the various metrics. \\
\citet{DBLP:journals/nle/GrahamBMZ17_crowd_alone} critically examine whether the assessment of the MT evaluation metrics is robust and if monolingual evaluators can be relied upon for human evaluations as opposed to experts or biligual evaluators. 
\cite{DBLP:journals/csl/DusekNR20_e2enlg} note the highly varied ranking outcomes while using automatic versus human evaluations. They particularly observe that with most automatic metrics, the fluency of the language can be reliably determined and that seq2seq models perform well on fluency. However in terms of adequacy and correctness, on which the automatic metrics do not agree with human evaluations, seq2seq NLG models perform worse than rule-based models, but they still manage to procure high scores on the automatic metrics. 
\cite{DBLP:journals/corr/abs-1807-02202_chaganty} also report the effects of having good quality (low variance) human judgements on the correlations of the automatic metrics to optimize evaluation costs. 


\subsection{Methods to Evaluate Robustness of Metrics}
Automatic metrics have been analysed for adversarial scenarios or corner cases in which they fail. This is important to understand the reliability and robustness of the metrics, and to make progress in improving them. In this section we briefly discuss some of the techniques used to construct adversarial examples. Most of these are automated methods based on text manipulation.  
One of the earliest works in this direction \cite{callison-burch-etal-2006-evaluating_bleu_in_mt} demonstrated a permutation attack on BLEU. For instance, by splitting the translation at bigram mismatch points, the resulting units can be reordered without a causing a change to the BLEU2 score. That is, while the permuted sentence need not necessarily make sense, BLUE assigns the same score to it as long as the number of matching n-grams doesn't change. 
More recently, \citet{hodosh-hockenmaier-2016-focused} provide a framework to create incorrect captions by replacing people (or the scene) in the caption with other people (or scenes), or switching the roles of people in the same caption, or even using other captions from the dataset that share people or scene with a given caption. To check for robustness of the image caption evaluation metrics, \citet{DBLP:conf/eacl/KilickayaEIE17_wmd_ic} adopt this method to create distracted versions of image captions/descriptions for a sample image. They check whether the metrics correctly identify a changed scene or person. 
\citet{DBLP:conf/aaai/SaiGKS19Reeval} design a set of synthetic scenarios to check the sensitivity of trained evaluators on jumbling/reversing the word order, removing punctuation or stopwords, retaining or removing specific combinations of the parts of speech, etc. \citet{li2017adversarial} compute `Evaluator Reliability Error' (ERE) on trained metrics based on their performance in simplified scenarios including passing the ground-truth as both the hypothesis and reference, similarly passing the candidate sentence as both, and passing random unrelated text as hypothesis. 
\\
Another approach to create hard or adversarial examples is to check performance on other related tasks or sub-tasks.  
The authors of BERTscore \cite{bertscore} evaluate the robustness of different metrics by considering their performance on a parallel task: adversarial paraphrase detection. Since most metrics rely upon comparison of the hypothesis with the reference, it is useful to determine if the metrics can assign lower scores to these adversarial paraphrases as opposed to the corresponding original paraphrases. 


\section{Recommendations (possible future research directions)}
Based on the above survey of the field, we would like to make the following recommendations: \\

\noindent \textbf{Developing a common code base for evaluation metrics:} As the number of metrics being proposed continues to grow, it is very important that we, as a community, develop a common code base for these metrics. Such a common code base would have multiple benefits: (i) researchers would be able to easily evaluate their NLG systems using a much wider range of metrics as opposed to relying on a very few metrics whose code is easily available (e.g., most MT papers report only BLEU scores even though several studies show that it does not correlate well with human judgements) (ii) results reported across papers will be comparable and not differ due to implementation issues (iii) researchers proposing new evaluation metrics will be able to compare easily against existing metrics while not being bothered by implementation issues (currently, many new metrics tends to compare their results with only those metrics whose code is easily available) and (iv) researchers will be able to critically examine existing metrics if the code is readily available leading to a more careful scrutiny of these metrics (e.g., they could perform white-box attacks or evaluate the metrics using carefully crafted adversarial examples \cite{DBLP:conf/aaai/SaiGKS19Reeval}).

\noindent \textbf{Building datasets containing human judgements:} Development of automatic evaluation metrics relies on the availability of datasets containing tuples of the following form: context, reference response, proposed response and human scores for the proposed response. The proposed response could either be generated by a human, a synthetically created variant of a reference response (e.g., dropping stop words or replacing synonyms, etc) or generated by an NLG system. Each of these proposed responses would then be evaluated by a human and assigned a score(say, on a scale of 0 to 1). Such a dataset would thus contain a mix of good and bad responses with human scores for them. This could act as data for training and evaluating automatic metrics for the given task. Further, given the varied criteria which need to be satisfied across different tasks (as discussed in section \ref{sec:human_eval}) such datasets should contain multiple scores wherein each score corresponds to a specific criteria (fluency, adequacy, coherence, \textit{etc}). Despite the evolving interest in this field, there is still a scarcity of such datasets for multiple tasks and languages. The shared task on evaluation metrics at the annual WMT conference \cite{DBLP:conf/wmt/BojarGKS16_wmt-16-results,DBLP:conf/wmt/BojarGK17_wmt-17-results,wmt-2018-results,DBLP:conf/wmt/MaWBG19} is a good example of creating such standardised datasets. Such shared tasks with standardised datasets for multiple tasks and languages would enable rapid progress in this field.

\noindent \textbf{Developing task-specific context-dependent metrics:} As is evident from the taxonomy diagram, Figure \ref{fig:taxo}, 
most of the existing metrics are context-free and thus task independent (although it is possible that the metrics were proposed for a specific task but given that they ignore the context, they can be easily adopted for most NLG tasks). Even the context-dependent metrics have largely focused on the task of dialog evaluation. This is clearly a problem, as context plays a very important role in evaluating whether the generated hypothesis is correct. As illustrated with the help of examples, in section \ref{sec:nlg_tasks} and \ref{sec:human_eval}, context is required to ensure that the generated hypothesis is coherent, factually consistent and relevant to the context. In the absence of context, an evaluation metric can only check for the similarity between the given set of references and the generated hypothesis. This is inadequate for tasks such as dialogues generation, summarization, question answering, data2text generation, \textit{etc.} where a wide variety of responses are possible and word/semantic overlap with the set of references is neither sufficient nor necessary. 

\noindent \textbf{Developing more interpretable metrics:} This goes hand in hand with the two recommendation made above (i) collecting criteria specific human judgements (fluency, adequacy, coherence, informativeness, etc) and (ii) developing task specific metrics. Most existing metrics assign a single score to the hypothesis which is clearly in contrast to how human evaluations are done. In particular, humans are asked to assign multiple scores to a hypothesis with each score corresponding to a specific criteria. It makes sense to have a similar expectation from an automated evaluation metric. This is simply because a single score provided by current metrics is often not actionable. For example, if the overall BLEU score of a MT system is 0.3 then should the developer of the system work on improving the fluency of the system (language model) or the adequacy of the system (translation model) or both. Contrast this with an evaluation system which assigns a separate score for fluency and adequacy. The output of such an evaluation system would be more interpretable and hence provide clear future directions to the researchers and/or developers of the NLG system. This would also help end-users or reviewers judge which system is better. For example, while comparing two question generation systems one might prefer a system which scores low on fluency but high on answerability as this would mean that while the questions generated by the system are not  grammatically perfect, they still convey the essence of the question (e.g., ``who director of Titanic''?). However, in the absence of such fine-grained scores it is hard to interpret and compare the relative strengths/shortcomings of multiple systems. Some very recent works in the direction of interpretability of metrics are proposed in \cite{DBLP:journals/corr/abs-2008-08896_decomposable_amr,DBLP:journals/corr/abs-2008-10427_probe_tasks_metrics}.

\noindent \textbf{Creating robust benchmarks for evaluating evaluation metrics:} While some of the early metrics, such as BLEU, NIST, METEOR, ROUGE, \textit{etc.}, have been critically examined across a wide variety of tasks \cite{callison-burch-etal-2006-evaluating_bleu_in_mt, DBLP:conf/emnlp/NemaK18_qbleu, DBLP:conf/aaai/SaiGKS19Reeval, DBLP:conf/emnlp/LiuLSNCP16HowNot}, many of the recently proposed evaluation metrics have not yet been examined critically. To facilitate such studies, there should be focus on creating adversarial evaluation benchmarks which critically examine the robustness of these metrics. For example, one could create datasets which contain adversarially crafted responses for a given context (say, a summary which has a high word/entity overlap with the passage but is still irrelevant or factually incorrect). Such adversarial evaluation has helped in identifying gaps in other NLP tasks such as QA \cite{DBLP:conf/emnlp/JiaL17_adversarial_squad} and has also shown promise in identifying shortcomings of dialog evaluation metrics \cite{DBLP:conf/aaai/SaiGKS19Reeval}. In addition to adversarial evaluations, such studies should also focus on identifying specific biases in the proposed evaluation metrics. For example, GAN based evaluators \cite{li2017adversarial} are biased towards systems that they have been trained on. Similarly, it is possible that some dialog evaluation metrics have certain biases induced by the data they are pretrained on. For example, if a metric is pretrained on Reddit or social media conversations then it may be biased against more formally written conversations (which have a slightly different word usage and syntactic structure). Lastly, it is also important to closely examine the evaluation measures such as correlation which are used for evaluating such evaluation metrics as recent studies have shown that such measures can be unreliable in certain situations \cite{DBLP:journals/corr/abs-2006-06264_tangled_up_in_bleu}.

\section{Conclusion}
NLG is a rapidly evolving field with multiple tasks, datasets and methods being proposed at a rapid pace. While there is not doubt that the community has made significant progress in the last 70 years, there is still a need for robust automated evaluation metrics which can help us accurately quantify the progress made in the field. Developing such evaluation metrics is a challenging task given the wide variety of criteria that need to be checked while evaluating a generated hypothesis. Over the past few years, many evaluation metrics have been proposed: some task agnostic and others task specific. In this survey, we reviewed these metrics by organising them in a coherent taxonomy. In particular, we propose that at a high level the metrics can be categorised as context-free and context-dependent metrics. Within each of these categories there are trained and untrained metrics which rely on word based, character based or embedding based information to evaluate a hypothesis. This arrangement of existing metrics in the proposed taxonomy clearly shows that there is still a need for developing task-specific context-dependent metrics as most of the current metrics are context-free. This is a major gap in existing works as in many NLG tasks context plays a very important role and ignoring it is not prudent.

In addition to reviewing existing evaluation metrics, we also presented the different evaluation measures (Pearson correlation, Spearman's correlation and Kendall's Tau) and significance tests used for evaluating such metrics. We emphasised that in certain situations these measures may be unreliable and hence it is important to use the right metric (for example, Pearson correlation is appropriate mostly when we expect a linear relationship between the two variables). We also discussed several studies which criticise the use of existing evaluation metrics due to their  (i) poor correlation with human judgements (ii) uninterpretability (iii) inherent biases (iv) poor adaptability across tasks and (v) inability to capture task-specific nuances. Lastly, based on our extensive survey we provide a list of recommendations or possible future directions. In particular, we emphasize on the need for (i) developing a common code base for reproducible research and wider adoption of metrics (ii) building more task-specific datasets containing fine-grained human judgements (iii) building task-specific context-dependent metrics using such datasets (iv) developing more interpretable scores which can provide precise directions for improvement and (v) creating robust benchmarks for critically examining proposed metrics to reveal their shortcomings and biases.

\begin{acks}
We would like to thank Department of Computer Science and Engineering, IIT Madras and Robert Bosch Center for Data Sciences and Artificial Intelligence, IIT Madras (RBC-DSAI) for providing us resources required to carry out this research. We would also like to thank Google for supporting Ananya Sai through their Google India Ph.D. Fellowship Program. We thank Juri Opitz, Nikita Moghe, Pritha Ganguly, and Tarun Kumar for their helpful comments on the paper.
\end{acks}

\bibliographystyle{ACM-Reference-Format}
\bibliography{main}
\end{document}